\documentclass{article}

 \usepackage[nonatbib, preprint]{neurips_2020}

\usepackage[utf8]{inputenc} 
\usepackage[T1]{fontenc}    
\usepackage{hyperref}       
\usepackage{url}            
\usepackage{booktabs}       
\usepackage{amsfonts}       
\usepackage{nicefrac}       
\usepackage{microtype}      
\usepackage[table]{xcolor}
\usepackage{xcolor}
\usepackage{epsfig}
\usepackage{graphicx}
\usepackage{amsmath}
\usepackage{amssymb}
\usepackage{float}                  
\usepackage{multirow,bigdelim} 
\usepackage{bm}
\usepackage{nicefrac}
\usepackage{booktabs}
\usepackage{nicefrac}
\usepackage[table]{xcolor}
\usepackage{subcaption}
\usepackage{footnote}
\usepackage[subtle]{savetrees}  
\DeclareUnicodeCharacter{2212}{-}
\usepackage{algorithm}
\usepackage[noend]{algpseudocode}
\usepackage{chngcntr}

\colorlet{green}{black} 

\newcommand{\argmax}{\mathop{\mathrm{arg max}}}

\begin{document}
\title{Foveation for Segmentation of \\ Ultra-High Resolution Images }

\author{%
  Chen Jin$^{1}$\thanks{These authors contributed equally.}\; , Ryutaro Tanno$^{1,2\ast}$ \\  
  \textbf{Moucheng Xu}$^{1}$, \textbf{Thomy Mertzanidou}$^{1}$, \textbf{Daniel C. Alexander}$^{1}$ \\ \\
  $^1$ Centre for Medical Image Computing, University College London, UK \\
  $^2$ Healthcare Intelligence, Microsoft Research Cambridge, UK \\
  \texttt{chen.jin@ucl.ac.uk}, \quad \texttt{rytanno@microsoft.com} \\
}

\maketitle              
\begin{abstract}
\vspace{-2mm}
Segmentation of ultra-high resolution images is challenging because of their enormous size, consisting of millions or even billions of pixels. Typical solutions include dividing input images into patches of fixed size and/or down-sampling to meet memory constraints. Such operations incur information loss in the field-of-view (FoV) i.e., spatial coverage and the image resolution. The impact on segmentation performance is, however, as yet understudied. In this work, we start with a motivational experiment which demonstrates that the trade-off between FoV and resolution affects the segmentation performance on ultra-high resolution images---and furthermore, its influence also varies spatially according to the local patterns in different areas. We then introduce \textit{foveation module}, a learnable ``dataloader'' which, for a given ultra-high resolution image, adaptively chooses the appropriate configuration (FoV/resolution trade-off) of the input patch to feed to the downstream segmentation model at each spatial location of the image. The foveation module is jointly trained with the segmentation network to maximise the task performance. We demonstrate on three publicly available high-resolution image datasets that the foveation module consistently improves segmentation performance over the cases trained with patches of fixed FoV/resolution trade-off. Our approach achieves the SoTA performance on the DeepGlobe aerial image dataset. On the Gleason2019 histopathology dataset, our model achieves better segmentation accuracy for the two most clinically important and ambiguous classes (Gleason Grade 3 and 4) than the top performers in the challenge by 13.1\% and 7.5\%, and improves on the average performance of 6 human experts by 6.5\% and 7.5\%. Our code and trained models are available at \url{https://github.com/lxasqjc/Foveation-Segmentation}.


\end{abstract}

\vspace{-4mm}
\section{Introduction}
\begin{figure}[htbp]
\centering
\vspace{-7mm}
\includegraphics[width=0.9
\textwidth]{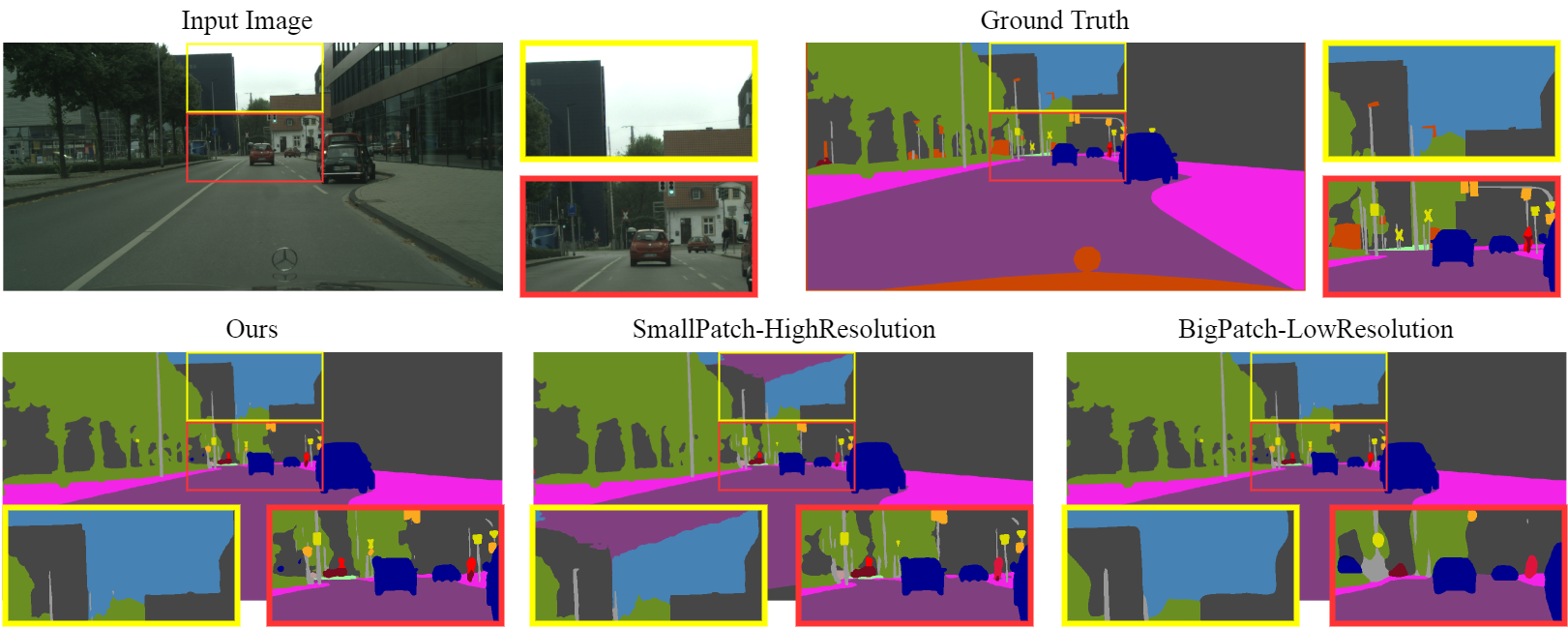}

\caption{\footnotesize A validation segmentation example on CityScapes\cite{cordts2016cityscapes} under two trade-off configurations: BigPatch-LowResolution case using patch size of $2048\times1024$ pixels and downsampling rate of 0.5 to original resolution; SmallPatch-HighResolution case using patch size of $1024\times512$ pixels and no downsampling applied. Yellow and red boxes highlight sub-areas containing coarse and fine scale context. The results show that our approach combines the strength of both BigPatch on capturing coarse context and SmallPatch on capturing fine context.} \label{fig:problem_highlight}
\vspace{-7mm}
\end{figure}
\vspace{-4mm}

Today, ultra-high resolution images are widely accessible to the computer vision community and therefore there is an increasing demand for effective analysis. Such images are commonly used in an array of scientific tasks, such as urban scene analysis with image size of $2048\times1024$ \cite{cordts2016cityscapes}, geospatial analysis with image size of $5000\times5000$ \cite{maggiori2017can} and histopathological images with image sizes up to $10,000\times10,000$ pixels \cite{srinidhi2019deep}. In particular, semantic segmentation, a problem of assigning each pixel to different semantic categories, forms an integral component of such analysis as it allows for a granular understanding of the scenes. However, manual analysis of ultra-high resolution images is extremely time-consuming, and is prone to false negative predictions due to the enormous image size, motivating the development of automated methods. 

Deep Learning (DL) based methods have been recently adopted to improve the segmentation of high-resolution images. However, due to the high GPU memory requirements, modern DL models with sufficient complexity cannot operate on the whole ultra-high resolution images in many practical settings. To mitigate this issue, the images are typically dissected into smaller patches and/or down-sampled to fit into the available GPU memory \cite{srinidhi2019deep, zhao2017pyramid, zhao2017self}. To exploit all the available GPU memory thus requires one to trade-off the field of view (FoV) i.e., spatial extent of context, against the spatial resolutions i.e., level of image detail. Tuning this trade-off exhaustively is expensive \cite{seth2019automated}, and as a result, it is commonly set based on crude developers' intuitions. Moreover, as \cite{seth2019automated} points out, the optimal trade-off is application or even image dependent (e.g., some parts of the images may require more context than local details), and thus the existence of an ``one-size-fits-all'' FoV/resolution trade-off for a given application is highly questionable.  

A considerable amount of work has attempted to alleviate the issue of subjectivity in tuning this trade-off by learning to merge multi-scale information, in both computer vision \cite{chen2016attention,chen2019collaborative} and medical imaging \cite{kamnitsas2017efficient}. Specifically, these approaches learn representations from multiple parallel networks and then aggregate information across different scales before making the final prediction. DeepMedic\cite{kamnitsas2017efficient} is one pioneering example in this category, which consists of two parallel networks processing input patches of different scales, and addresses the task of lesion segmentation in voluminous multi-modal 3D MRI data. Another example is \cite{li2019classification}, which is designed to work specifically with mega-pixel histopathology images. In computer vision, the authors of \cite{chen2019collaborative} proposed a method for integrating multi-scale information by enforcing global and local streams to exchange information with each other, and attained the state-of-the-art performance on the segmentation of ultra-high resolution satellite images. Another notable approach \cite{chen2016attention} uses the attention mechanism to softly weight the multi-scale features at each pixel location, and demonstrate impressive performance on multiple challenging segmentation benchmarks. However, these approaches share some limitations: 1) presence of multiple parallel networks, which can be computationally expensive; 2) manual selection of a limited number (2 or 3) of scales; 3) reliance on specific choices of neural network architecture.

In this work, we first demonstrate empirically on three public segmentation datasets of ultra-high resolution images that the choice of the input patch configuration (i.e., FoV/resolution trade-off) indeed considerably influences the segmentation performance. Secondly, motivated by this finding, we then propose \textit{foveation module}, a data-driven “data loader” that learns to provide the segmentation network with the most informative patch configuration for each location in an ultra-high-resolution image. Specifically, the foveation module processes a downsampled version of the given ultra high-resolution image and estimates the distributions over multiple patches of different resolution/FoV trade-offs at respective locations. This hierarchical approach to segmentation is inspired by the way in which human annotators provide segmentation labels to an extremely large image --- they saccade their gaze through the whole image and zoom in at different locations with the appropriate amount in order to acquire the required local and contextual information. 

The foveation module can be trained jointly in an end-to-end fashion with the segmentation network to optimise the task performance.  Additionally, our foveation module, as a learnable/adaptive data-loader, can be used to augment a wide range of existing segmentation architectures (as we show empirically). We demonstrate the general utility of our approach using three public datasets from different domains (Cityscape, DeepGlobe and Gleason2019 challenge dataset), where we show a light-weight implementation of the foveation module can boosts segmentation performance with little extra computational cost.

\section{Related Work}
\label{related_work}
\vspace{-3mm}
\subsection{Multiscale Architectures}
\vspace{-3mm}

Multi-scale approaches \cite{chen2014semantic, chen2016attention, hariharan2015hypercolumns} have achieved good performances in segmentation task by either aggregating features at different layers or features of multiple inputs with different resolutions. Such aggregation process is performed in either sequential or parallel fashion. A typical sequential approach such as RefineNet \cite{lin2017refinenet} fuses features from multiple inputs of different resolutions progressively starting from the lowest. Feature fusion can also be done in parallel. For example, U-net \cite{ronneberger2015u} directly concatenates the low-level features from the encoder to the high-level features in the decoder via skip connections. The contracting-expanding variants of U-net is very popular \cite{noh2015learning, badrinarayanan2017segnet, zhao2017pyramid}. A similar approach, called Feature Pyramid Network (FPN) \cite{lin2017feature} fuses the features in a hiearachical fashion as in U-net, but makes predictions at respective feature levels and aggregates them into the final output.

Recently, the high-resolution network (HRNet) \cite{sun2019high} was proposed to combine the merits of sequential and parallel approaches and achieved the state-of-the-art performance on the CityScapes segmentation dataset \cite{cordts2016cityscapes}. HRNet merges features at different scales in parallel, and then repeats this merge in a sequential manner from  higher resolution level. Another notable approach, specially designed to segment ultra-high resolutions images is GlobalLocalNet(GLNet) \cite{chen2019collaborative}, where the low-level features from the massively downsampled input image and the high-level features from local patches of the original resolution are aggregated in a spatially consistent fashion.

While the above approaches have demonstrated considerable performance, they still resort to the specific designs of network architectures. In contrast, our approach can be used on a wide range of network --- we will demonstrate later the benefits of augmenting UPerNet \cite{xiao2018unified} (an extension of FPN \cite{lin2017feature}) and HRNet \cite{sun2019high}. Furthermore, many of the above multi-scale methods use manually tuned combination of the crop size (i.e., FoV) and resolutions of the input patches, or use a set of multiple randomly chosen such combinations \cite{zhao2017pyramid, zhao2017self, sun2019high} to ensure good performance. Here we provide a systematic analysis to confirm the considerable influence of these two factors on the segmentation performance, and present a mechanism (foveation module) to infer the optimal patch configurations at different image locations.  

\vspace{-3mm}

\subsection{Hierarchical vision systems}
\vspace{-3mm}

Region-proposal approaches for object detection tasks have shown that optimal region crops vary spatially depending on local patterns. A typical region-proposal framework consists of a hierarchy of tasks of 1) proposing region crops and 2) the downstream classification \cite{girshick2014rich, girshick2015fast} or segmentation task of the cropped region \cite{he2017mask}. Our approach shares a similar structure but differs in that the first task proposes a set of desirable input patch configuration (FoV/resolution-tradeoff) at different image locations to maximize downstream performance. Also our approach is end-to-end while typical region-proposal approaches train the two hierarchical tasks separately for computational efficiency. The approach that is closest to ours is ``Learning to Zoom'' in \cite{recasens2018learning} where they introduce a learnable module to change the resolution of the input images in a spatially varying way to emphasise the salient parts of the data. However, such approach assumes that the task network still operates on the whole downsampled image, limiting its applicability to ultra-high resolution images.

\vspace{-3mm}
\subsection{Processing ultra-high resolution images in computer vision}
\vspace{-3mm}

Most CNN-based approaches do not process images with size beyond $512^2$ in 2D or $64^3$ in 3D \cite{girdhar2016learning, choy20163d}. Non-uniform grid representation is becoming a common approach in computer vision to alleviate the memory cost by modifying input images in a non-uniform fashion to improve efficiency, such as meshes \cite{wang2018pixel2mesh, gkioxari2019mesh}, signed distance functions \cite{mescheder2019occupancy}, and octrees \cite{tatarchenko2017octree}. Recently, Marin et al. \cite{marin2019efficient} proposed an efficient semantic segmentation network based on non-uniform subsampling of the input image prior to processing with a segmentation network. However, these lines of work do not consider optimising such representations to maximise the downstream task of interest. A related but alternative strategy is proposed in \cite{katharopoulos2019processing} in which attention weights are learned over the input image to sample a small informative subset for the downstream classification task. A similar approach was introduced in \cite{shen2019globally} but based on the use of saliency maps for the task of breast cancer screening with high-resolution mammography images.  However, these works only addressed the problem of classification and is not directly applicable to the segmentation task which requires all locations to be examined.


\vspace{-1mm}
\section{Methods} 

\vspace{-3mm}
In this section, we first perform a motivation experiment to illustrate the impact of the patch FoV/resolution trade-off on the segmentation performance and its spatial variation across the image. Motivated by this finding, we then propose \textit{foveation module}, a module that learns to provide the segmentation network with the most informative patch configuration for each location in a ultra-high-resolution image. 

\vspace{-2mm}
\subsection{Patch Configuration Matters in Segmentation}

\label{empirical_motivation}

The first part of our work performs an empirical analysis to investigate a key question: ``How does the FoV/resolution of training input patches affect the final segmentation performance?''. To this end, we use the following three ultra-high-resolution segmentation datasets  from different domains.

\begin{figure}[h]
\centering
\includegraphics[width=1.0\textwidth]{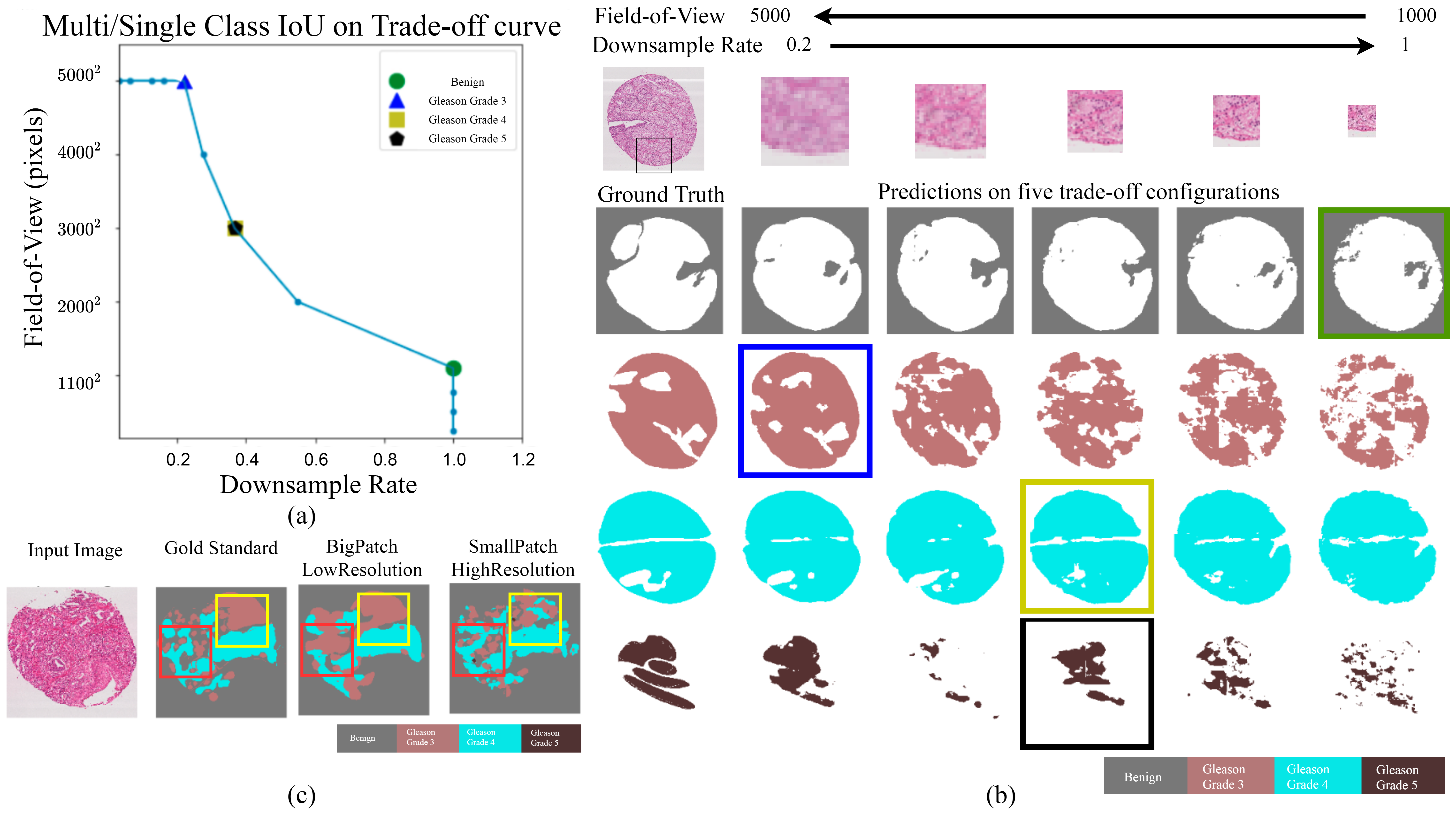}
\caption{\footnotesize Quantitative (a) and qualitative evidence that the optimal FoV/resolution trade-off varies over classes (b) and locations (c). (a): segmentation networks trained with different trade-off configurations where best performance for each class is highlighted (see Legends for references). (b): FoV decreases from left to right; Downsampling rate increases from left to right. In each row, the best segmentation result among the five trade-off configurations is highlighted, which corresponds to the trade-off configuration in (a). (c): An example under two trade-off configurations: BigPatch-LowResolution case uses patch size of $5000^2$ pixels and downsampling rate of 0.2 to original resolution; SmallPatch-HighResolution case uses patch size of $2000^2$ pixels and downsampling rate of 0.5 to original resolution.}
\label{fig:motivation}
\end{figure}

\begin{table}[H]
  \footnotesize
  \caption{\small Dataset summary. More details in Supplementary Material}
  \label{Key_dataset_elements}
  \centering
  \begin{tabular}{lllll}
    \toprule
    Dataset & Content & Resolution (pixels) & Number of Classes \\
    \midrule
    Cityscape \cite{cordts2016cityscapes} & urban scenes &  $2048 \times 1024$ & 19  \\
    DeepGlobe \cite{DeepGlobe18} & aerial scenes &  $2448 \times 2448$ & 6 \\
    Gleason2019 \cite{Gleason2019} & histopathological &  $5000 \times 5000$ & 4 \\
    \bottomrule
  \end{tabular}

\end{table}
In the interest of space,  we only show results on the Gleason 2019 histopathology dataset (see the supplementary material for similar results on DeepGlobe\cite{DeepGlobe18} and Cityscape\cite{cordts2016cityscapes}). We first train a set of different segmentation networks, each with a different combination of FoV and downsampling rate (i.e., resolution); see the small blue dots on the curve in Fig.~\ref{fig:motivation}(a)). We note that the maximum tensor size of the input patch is capped at $1000^2$ and constant along the curve in Fig.~\ref{fig:motivation}(a). The segmentation network with the best performance for each class is highlighted in different shaped marks (see Fig.~\ref{fig:motivation}(a)). It is clear that there is no ``one-size-fits-all'' patch configuration of the training data that leads to the best performance overall and for all individual classes. Fig.~\ref{fig:motivation}(b) illustrates visually the variation of the segmentation performance. In addition, even within each class, we find that the optimal patch configuration can vary between different spatial locations as shown in Fig.~\ref{fig:motivation}(c). These observations altogether imply that the standard patch sampling scheme with a pre-set FoV/Resolution trade-off is sub-optimal, highlighting the potential benefits of a more intelligent strategy that can adaptively select the patch of the most informative configuration to describe the local patterns at a given location.

\vspace{-2mm}
\subsection{Foveation Module for Adaptive Patch Configuration}\label{sec:Foveation_method}
\renewcommand{\thefootnote}{1}

Motivated by the above finding, we introduce \textit{foveation module}, a data-driven patch sampling strategy which selects, at each spatial location in an ultra-high resolution image, an appropriate configuration (i.e., resolution and FoV) of the local patch that feeds into the segmentation network. The inspiration for our method roots from the ways in which human experts segment high-resolution images --- starting from a low-resolution bird's-eye view of the whole image \footnote{Screen display or human vision typically have lower resolutions than that of the ultra-high resolution images of interest in this work.}, the annotators navigate their gaze through different locations and zoom in to the right extent to collect both local and contextual information. The magnification scale is controlled by, what is called, \textit{foveation} (i.e., the process of adjusting the focal length of the eye, the distance between the lens and fovea). The proposed adaptive patch-selection scheme is akin to this process, and hence the name, foveation module. Foveation module can be seen as a learnable ``dataloader'' that is optimised to maximise the performance of a given segmentation network.

\begin{figure}[h]
\centering
\includegraphics[width=1.0\linewidth]{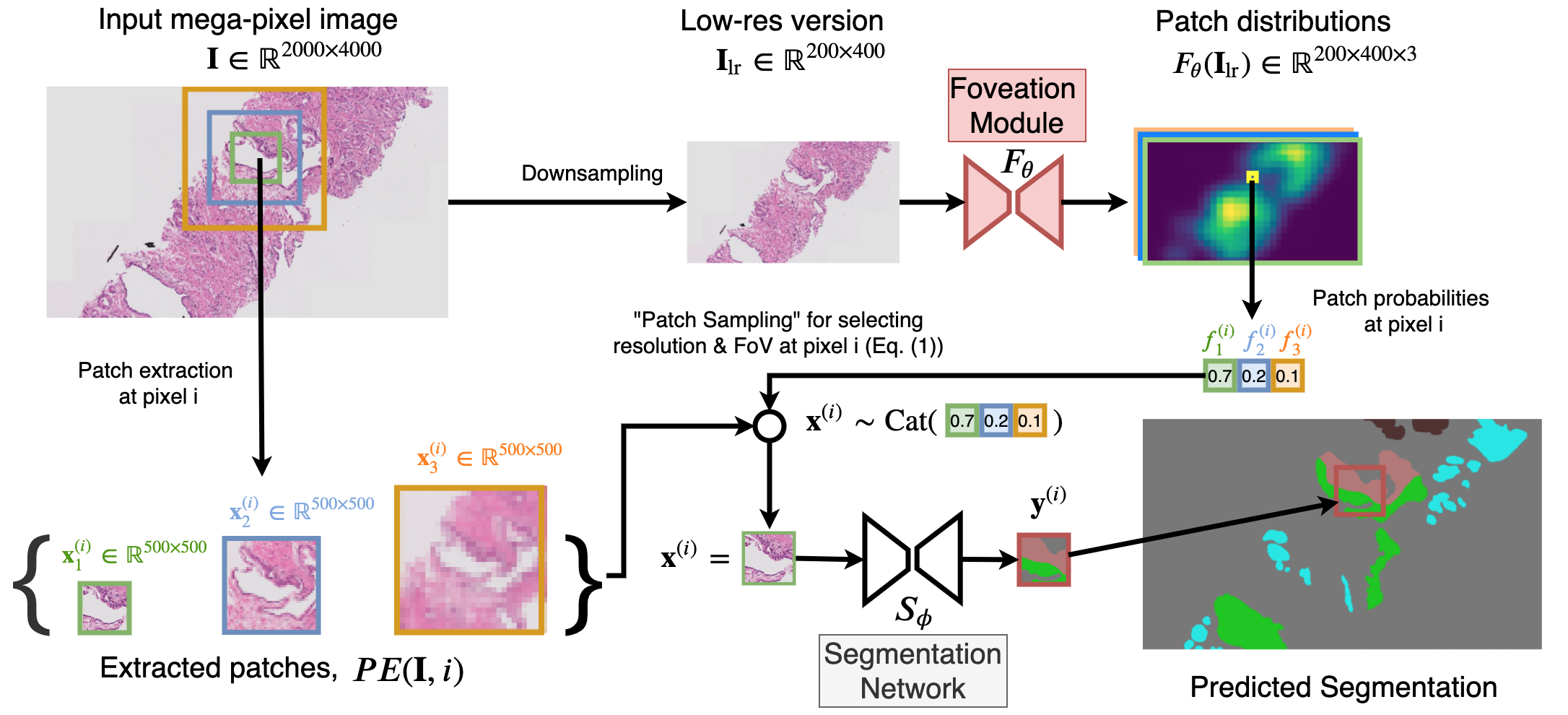}
\vspace{-1mm}
\caption{\footnotesize Architecture schematic. } 
\label{fig:architecture}
\end{figure}

Fig.\ref{fig:architecture} provides a schematic of the proposed method. \textit{Foveation module} takes a low-resolution version of a mega-pixel input image and generates importance weights over a set of patches with varying spatial FoV/resolution at different pixel locations. Then, the \textit{segmentation network} processes the input patches based on the outputs of the foveation module, and estimates the corresponding segmentation probabilities. Note that the choice of this segmentation network only requires them to operate on a single input patch/image---we will later demonstrate this by augmenting two recent and different architectures, namely UPerNet \cite{xiao2018unified} (Pyramid Pooling + FPN head) and HRNet \cite{sun2019high}.

More specifically, for each mega-pixel image $\mathbf{I} \in \mathbb{R}^{H\times W \times C}$ where H, W, C denote the height, width and channels respectively, we compute its lower resolution version $\mathbf{I}_{\text{lr}} \in \mathbb{R}^{h\times w\times C}$. We also define a ``patch-extractor'' function
$PE(\textbf{I}, i)=\{\mathbf{x}_{1}^{(i)}(\textbf{I}),...,\mathbf{x}^{(i)}_{D}(\textbf{I})\} $ that extracts a set of $D$ patches of varying field-of-view/resolution (but the same number of pixels) from the full resolution image $\textbf{I}$ centered at the corresponding $i^{\text{th}}$ pixel in $\mathbf{I}_{\text{lr}}$ (see Fig.~\ref{fig:architecture} for a set of examples). \textit{Foveation module}, $F_{\theta}$, parametrised by $\theta$, takes the low-resolution image $\mathbf{I}_{\text{lr}}$ as the input and generates the probability distributions $F_{\theta}(\mathbf{I}_{\text{lr}}) \in [0, 1]^{h\times w\times D}$ over patches $PE(\textbf{I}, i)=\{\mathbf{x}_{1}^{(i)},...,\mathbf{x}^{(i)}_{D}\}$ at respective spatial locations $i\in \{1,...,wh\}$ in $\mathbf{I}_{\text{lr}}$. In other words, the values of $F_{\theta}(\mathbf{I}_{\text{lr}})$ at the $i^{\text{th}}$ pixel define a $D$-dimensional probability vector $f^{(i)}_{\theta}(\textbf{I}_\text{lr}) = [f_{\theta,1}^{(i)}(\textbf{I}_\text{lr}), ..., f_{\theta,D}^{(i)}(\textbf{I}_\text{lr})]$ over the extracted patches. We then select the input patch by sampling from this patch distribution: \begin{equation}\label{eq:discrete}
\mathbf{x}^{(i)} \sim \text{Categorical}\big{(}f^{(i)}_{\theta}(\textbf{I}_\text{lr})\big{)} 
\end{equation}
and feed it to the \textit{segmentation network}, $S_{\phi}\big{(}\mathbf{x}^{(i)}\big{)}$, parametrised by $\phi$ to estimate the corresponding segmentation probabilities. We note here that the spatial extent of the predicted segmentation corresponds to the area covered by the input patch with the smallest field of view in $PE(\textbf{I}, i) $. 

During training, we would like to encourage the foveation module to output meaningful probabilities over patches at different spatial locations. 
In particular, the patch probabilities are desired to reflect the relative ``informativeness'' of different patch configurations at each location for the downstream segmentation task. To this end, we jointly optimise the parameters $\{\theta, \phi\}$ of both the foveation module and the segmentation network to minimise the segmentation specific loss function, $\mathcal{L}(\mathbf{x}^{(i)}; \theta, \phi)$ (e.g., cross entropy + L2-weight-decay). However, due to the non-differentiable nature of patch sampling from discrete distributions, one cannot naively apply stochastic gradient descent. We devise and evaluate several solutions as detailed in Sec.~\ref{sec:learning}. We also note that, for computational efficiency, for each mega-pixel image $\mathbf{I}$, we randomly select a subset of pixels from its low resolution counterpart $\mathbf{I}_{\text{lr}}$, compute the corresponding input patches, feed them to the segmentation network and compute the losses. 

At inference time, we segment the whole mega-pixel image $\mathbf{x}$ by aggregating predictions $S_{\phi}\big{(}\mathbf{x}^{(i)}\big{)}$ at different locations $i$. To speed up the inference, we also optionally downsample $F_{\theta}(\mathbf{I}_{\text{lr}})$ to reduce the number of locations $i$ over which to sample patches as explained in detail in Sec.~\ref{sec:results}.

\subsection{Learning the Spatial Distribution of Patch Configurations}\label{sec:learning} 

The sampling of input patches from discrete distributions in eq.~\eqref{eq:discrete} creates discontinuities giving the objective function $\mathcal{L}(\mathbf{x}^{(i)}; \theta, \phi)$ zero gradient  with respect to $\theta$, the parameters of the foveation module. We devise the following approximations to address this problem. 

\paragraph{Gradient estimation with Gumbel-Softmax:} In this approach, we approximate each discrete distribution, $\text{Categorical}(f^{(i)}_{\theta}(\textbf{I}_\text{lr}))$ by the so-called Concrete \cite{maddison2016concrete} or Gumbel-Softmax distribution \cite{jang2016categorical}, denoted by $\text{GSM}(f^{(i)}_{\theta}(\textbf{I}_\text{lr}), \tau)$. GSM is a continuous relaxation which allows for patch sampling, differentiable with respect to the parameters $\theta$ of the foveation module through a reparametrisation trick. The temperature term $\tau$ adjusts the bias-variance trade-off of gradient approximation; as the value of $\tau$ approaches 0, samples from the GSM distribution become one-hot (i.e. lower bias) while the variance of the gradients increases. In practice, we start at a high $\tau$ and anneal to a small but non-zero value as in  \cite{jang2016categorical,gal2017concrete,bragman2019stochastic} as detailed in supplementary materials. 

We have also experimented with another popular gradient estimator, REINFORCE \cite{williams1992simple}. The early experiments suggested that the optimisation was challenging likely due to the well-known high variance of the estimator. One could also use more sophisticated, unbiased estimators with low-variance such as REBAR \cite{tucker2017rebar} and RELAX \cite{grathwohl2017backpropagation}, but the GSM-based approximation (although only unbiased in the limit of $\tau \longrightarrow 0$) worked well for the initial experiments in this paper.

\paragraph{Mean Approximation:} Instead of sampling patches at each pixel location $i$, we compute the average input patch weighted by the estimated probabilities and feed it to the segmentation network: 
\begin{equation}
    \mathbb{E}[\mathbf{x}^{(i)}] = \sum_{d=1}^{D} f_{d}^{(i)}(\textbf{I}_\text{lr})\cdot \mathbf{x}^{(i)}_{d}(\textbf{I})
    \label{eq:input_patch_1}
\end{equation}
Here $f_{d}^{(i)}(\textbf{I}_\text{lr})$ denotes the value of $F_{\theta}(\mathbf{x}_{\text{lr}})$ at $i^{\text{th}}$ pixel, and quantifies the ``importance'' of the $d^{\text{th}}$ patch at that location. With this approach, the objective function becomes fully differentiable with respect to $\theta$. Such mean approximations are commonly employed in the attention literature, and have shown efficacy in different contexts e.g., the multi-instance learning for ultra-high resolution images \cite{ilse2018attention} and the “deterministic soft” attention for image captioning in \cite{xu2015show}. 

\paragraph{Mode Approximation:} We also experiment with the option to select the most probable patch:
\begin{equation}
    \text{Mode}[\mathbf{x}^{(i)}] = \argmax_{\mathbf{x}^{(i)}_{d} \in  PE(\textbf{I}, i)}(f_{d}^{(i)}(\textbf{I}_\text{lr}))
    \label{eq:input_patch_2}
\end{equation}
Since the gradient is not well-defined in this case, we approximate it with a straight-through estimator \cite{bengio2013estimating} which directly copies the gradient from the preceding layer. Such an approach, while biased, has also been shown effective in learning discrete representations in VAE-type generative models of high-dimensional data \cite{van2017neural} in the presence of an $\argmax$ function.

\vspace{-2mm}


\section{Experiments and Results}\label{sec:results} 
\vspace{-2mm}
In this section, we evaluate the performance of our foveation approach on three datasets against baselines described in \ref{result:baselines}. We show that our foveation module can learn the spatial variation of optimal trade-offs in \ref{result:eval_foveation}, leading to better performance in \ref{result:performance}. Different variants of our method i.e., Gumbel-softmax, mean and mode approximations, are referred to as `Ours-GSM', `Ours-Mean' and `Ours-Mode' respectively. 

We employ the same training scheme unless otherwise stated. The \textit{foveation module} is defined as a small CNN architecture comprised of 3 convolution layers plus a softmax layer. The segmentation network was defined as a deep CNN architecture, with HRNetV2-W48\cite{sun2019high} used in Cityscape and Gleason2019 dataset, and UPerNet \cite{xiao2018unified} used in DeepGlobe dataset. In all settings, $PE(\mathbf{I},i)$ extracts, at each image location, a set of $5$ patches of varying FoV/resolution-tradeoffs. Full details are provided in Supplementary Material.

\vspace{-2mm}
\subsection{Baselines}
\vspace{-2mm}
\label{result:baselines}

We compare our method against a variety of baselines. Firstly, we consider the same segmentation networks (HRNet or UPerNet), but trained on input patches of fixed FoV/resolution-tradeoffs---there are five of them since we consider 5 different patch configurations. We also include the results from ensembling these five baselines (``Ensemble''). To further investigate the benefits of learning the probabilities over patch configurations, we also compare against the cases trained with randomly sampled input configuration (``Random'') or the average input patches with equal weights (``Average''). More details are in the Supplementary Material.

\subsection{Quantitative and Qualitative Comparison}
\label{result:performance}
\vspace{-2mm}

\begin{figure}[h]
\centering
\vspace{-2mm}
\includegraphics[width=\textwidth]{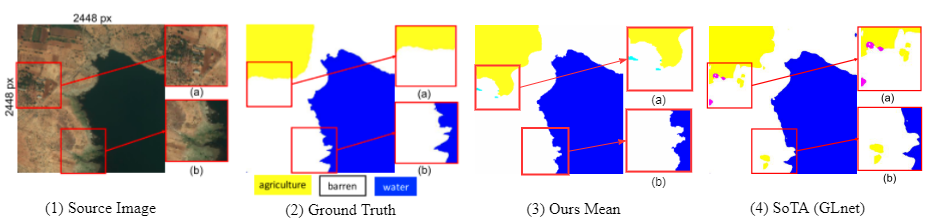}
\caption{\footnotesize Qualitative comparison of our method versus the SoTA results given in GLnet \cite{chen2019collaborative}, color code: yellow: Agriculture; pink: Rangeland; blue: Water; white: Barren.} \label{Deepglobe_Qualitative}
\end{figure}

\begin{figure}[h]
\centering
\vspace{-4mm}
\includegraphics[width=\textwidth]{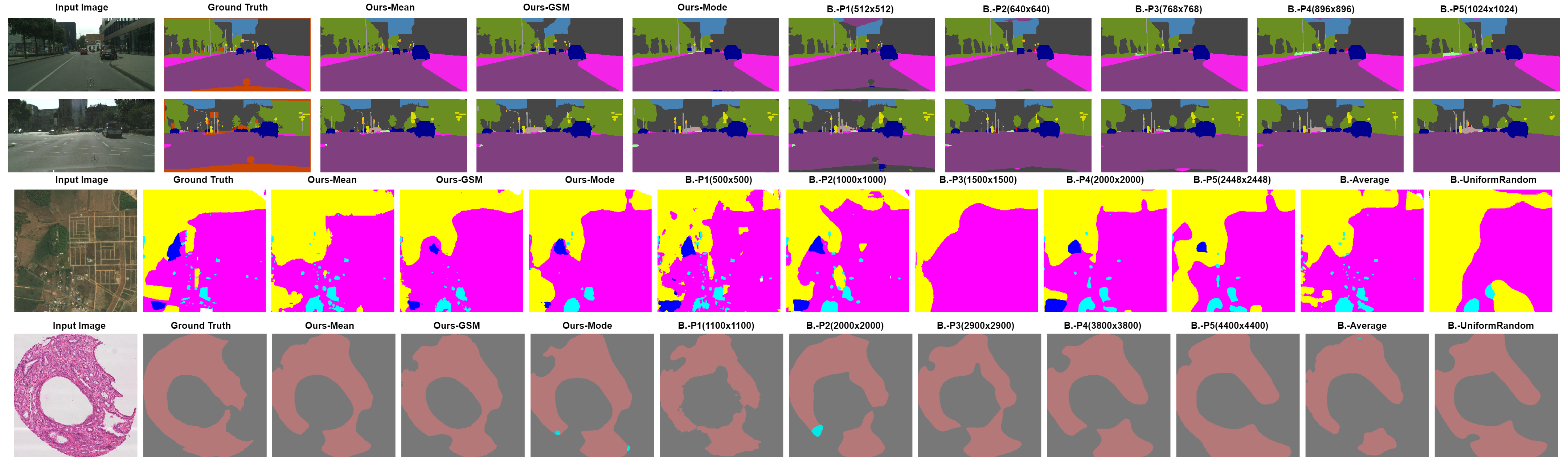}
\caption{\footnotesize Qualitative comparison of our method versus baselines, each row gives examples from different datasets (row 1-2 from Cityscape, row 3 from Deepglobe, row 4 from Gleason2019); each column gives input image, ground truth and sementation results. More examples are provided in Supplementary Material.} \label{Segmentation_main}
\end{figure}

The quantitative comparison between our methods and the baselines is given in Table~\ref{results: Performance_all_datasets}. We show segmentation performance for all classes in DeepGlobe and Gleason2019 dataset, while only all classes average is shown for Cityscape due to space limit (class-wise performance provided in Supplementary Material instead). Table~\ref{results: Performance_all_datasets} shows that our methods (especially `ours-Mean') generally shows improvement over the baselines, illustrating the benefits of using more desirable FoV/Resolution at each location. It's also worth noting that 6\% boost in mIoU is achieved for Cityscape. Fig.~\ref{Deepglobe_Qualitative} \&~\ref{Segmentation_main} illustrate these numerical improvements reflect meaningful differences in segmentation quality.

Our approach also achieves favourable results with respect to the published results. Firstly, Table~\ref{results: Performance_all_datasets} shows that our approach provides 2.4\% boost over comparable SoTA \cite{chen2019collaborative} on the DeepGlobe dataset with visually noticeable differences in segmentation quality as shown in Fig.~\ref{Deepglobe_Qualitative}), providing finer details in the coastal area and less miss-classification of agriculture in sub-area (b). Secondly, on Gleason2019, our model achieves better segmentation accuracy for the two most clinically important and ambiguous classes (Gleason Grade 3 and 4) than the top performers in the challenge by 13.1\% and 7.5\%, and improves on the average performance of 6 human experts by 6.5\% and 7.5\% (details are provided in Supplementary Material). 

Among the three different variants of our method, the mean approximation seems to be the most effective approach, achieving top performance for most single class cases. This is likely due to the optimisation since GSM and mode approximations rely gradient estimators. We do note, however, that in some cases, GSM or mode approximations perform better than the mean approximation (e.g. "Forest" and "Barren" in DeepGlobe of Table~\ref{results: Performance_all_datasets}). The third row in Fig.~\ref{Segmentation_main} is a visual example on this point.

\begin{table}[htbp]
  \centering
  \vspace{-3mm}
  \scriptsize
  
  \begin{tabular}{lllllllllllll}
    \toprule
    Dataset & \multicolumn{7}{c}{DeepGlobe} & \multicolumn{4}{c}{Gleason2019} & \multicolumn{1}{c}{City.} \\
    \cmidrule(lr){2-8} \cmidrule(lr){9-12} \cmidrule(lr){13-13}
    {Class} & All & U. & A. & R. & F. & W. & B. & All & Benign & Grade 3 & Grade 4 & All \\
    \midrule
    Baseline-Patch 1 & 73.8 & 79.4 & 88.2 & 45.1 & 77.2 & \cellcolor{red!15} {\bfseries87.4} &  65.4 & 69.3 & 81.0 & 61.8 & 65.0 & 70.2\\
    Baseline-Patch 2 & 73.4 & 79.0 & 88.4 & 43.6 & 77.7 & 85.6 & 66.2 & 70.1 & 80.5 & 64.4 & 65.3 & 70.6\\
    Baseline-Patch 3 & 56.4 & 57.0 & 81.4 & 27.8 & 62.3 & 69.1 & 40.7 & 70.7 & 81.0 & \cellcolor{blue!15} 64.9 & 66.1 & 69.8\\
    Baseline-Patch 4 & 72.3 & 77.2 & 87.8 & 41.8 & 78.5 & 84.2 & 64.2 & 67.3 & 76.4 & 61.4 & 64.0 & 68.0\\
    Baseline-Patch 5 & 71.3 & 76.9 & 87.0 & 39.6 & 79.1 & 83.0 & 62.2 & 61.4 & 67.5 & 55.6 & 61.0 & 67.7\\
    Baseline-Random & 46.4 & 56.3 & 77.0 & 20.1 & 41.0 & 45.8  &  38.5 & 64.9 & 72.2 & 58.6 & 63.8 & 50.3\\
    Baseline-Average & 73.5 & 78.4 & 88.1 & 45.0 &  \cellcolor{red!15} {\bfseries80.8} & 82.3 & 66.7 & 65.7 & 78.8 & 52.2 & 66.0 & 56.6\\
    Ensemble  & \cellcolor{red!15} {\bfseries74.3} &  \cellcolor{blue!15}79.4 &  \cellcolor{red!15} {\bfseries88.7} & 44.5 & 80.0 & 86.3 &  \cellcolor{red!15} {\bfseries66.8} & \cellcolor{blue!15} 71.0 & 80.8 & \cellcolor{red!15} {\bfseries65.4} & \cellcolor{blue!15} 66.9 & \cellcolor{blue!15}72.5\\
    \midrule
    Ours-Mean  & \cellcolor{blue!15} 74.0 &  \cellcolor{red!15}{\bfseries79.5} &  \cellcolor{red!15}{\bfseries88.7} &  \cellcolor{red!15}{\bfseries45.2} & 77.6 & \cellcolor{blue!15}86.4 & 66.6  &  \cellcolor{red!15}{\bfseries71.1} & \cellcolor{red!15} {\bfseries82.4} & 63.0 & \cellcolor{red!15} {\bfseries67.8} &  \cellcolor{red!15}{\bfseries76.1}\\
    Ours-Mode & 73.2 & 77.5 & 88.2 & 43.4 & 79.9 & 84.4 & 65.5 & 69.5 & \cellcolor{blue!15} 81.6 & 61.3 & 65.6 & 70.5\\
    Ours-GSM & 73.8 & 78.4 & 88.5 & \cellcolor{blue!15} 44.7 & \cellcolor{blue!15} 80.1 & 84.4  &  \cellcolor{red!15} {\bfseries66.8} & 67.4 & 79.4 & 58.3 & 64.4 & 71.4\\
    \midrule
    GLnet \cite{chen2019collaborative} & 71.6 &- & - &  - & - & -  & - & - & - & - & -& -\\
    Clinical experts \cite{Gleason2019} & - & - & - & - & - & - & - & 66.9 & 83.9 & 56.4 & 60.3 & -\\
    \bottomrule
  \end{tabular} 
\vspace{1mm}
    \caption{\footnotesize Segmentation performance measured in IoU/mIoU (\%) on DeepGlobe, Gleason2019, Cityscape (City.) datasets. The best and the second best results are shown in red and blue. In last two rows, we quote GLnet result for DeepGlobe dataset, and average of 6 expert annotation performance for Gleason2019 dataset as reference. For class names in the second row, All referred to all classes average, and the rest single class codes are as follows: DeepGlobe: U.-Urban, A.-Agriculture, R.-Rangeland, F.-Forest, W.-Water, B.-Barren} 

  \vspace{-1mm} 
  \label{results: Performance_all_datasets}
\end{table}

\vspace{-3mm}
\subsection{Evaluation of Foveation Module}
\vspace{-2mm}
\label{result:eval_foveation}
In this section, we demonstrate that our foveation module can effectively learn spatial distribution of the FoV/Resolution trade-off. To visualise the learnt trade-off at each location, we use the probability predicted from foveation module $f^{(i)}_{\theta}(\textbf{I}_\text{lr})$ at each $i^{\text{th}}$ pixel in $\mathbf{I}_{\text{lr}}$ to calculate the weighted average FoVs of the given set of patches, which we refer to as ``foveation map''. In order to measure the quality of such foveation maps, we define a `Gold Standard' by visualising the FoV of the most performant patch configuration amongst the baselines trained fixed patches. It is, however, worth noting that such gold standard does not equate to the true optimal patch configuration of the given segmentation model and the dataset since only a limited set of patch sizes are evaluated. Details are provided in Supplementary Materials.

\begin{figure}[h]
\centering
\vspace{-4mm}
\includegraphics[width=\textwidth]{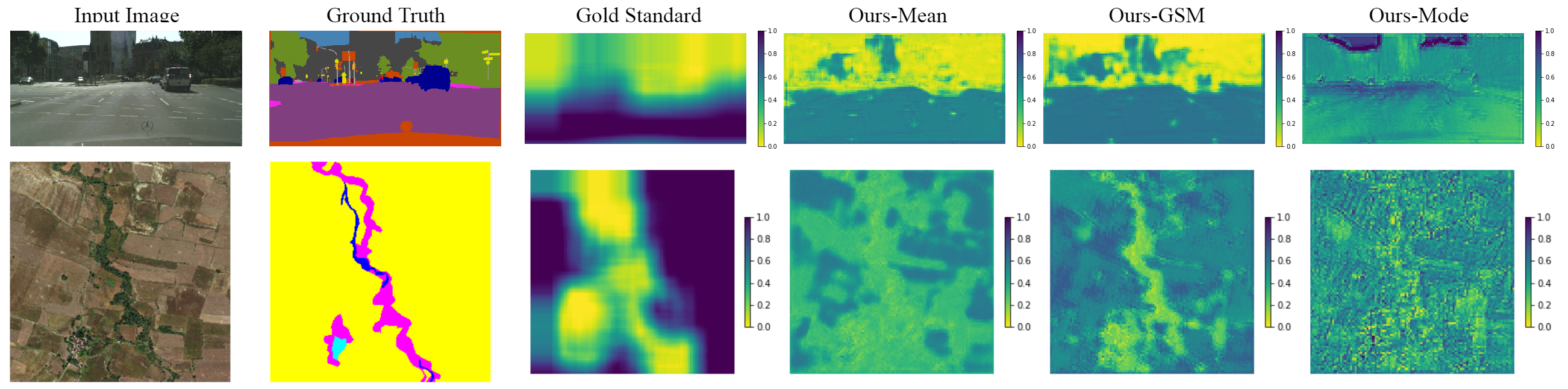}
\vspace{-2mm}
\caption{\footnotesize Qualitative results on the foveation maps. Examples from the Cityscape (1st row) and DeepGlobe (2nd row) are shown. 'Gold Standard' referred to the mIoU weighted FoV map, and 'Ours-Mean', 'Ours-GSM' and 'Ours-Mode' referred to as foveation output probability weighted FoV maps, i.e. foveation maps (FoVs are normalised between 0-1). More diverse examples are provided in Supplementary Material.}  
\label{Evaluate_foveation_main} 
\vspace{-2mm}
\end{figure}

\begin{table}[H]
\vspace{-1mm}
  \centering
   \caption{\footnotesize Mean Squared Error (MSE) between map of mIoU and foveation output weighted average FoV, with three ours approaches and over three datasets. Mean and standard deviation of MSE are calculated across all validation images in each dataset.}
  \footnotesize
  \vspace{1mm}
  \begin{tabular}{llll}
    \toprule
     & Ours-Mean Approximation & Ours-GSM & Ours-Mode Approximation \\
    \midrule
    Cityscape & \(0.084\pm0.043\) & \(0.082\pm0.038\) & \(0.184\pm0.048\) \\
    DeepGlobe & \(0.171\pm0.076\) & \(0.146\pm0.063\) & \(0.149\pm0.055\) \\
    Gleason2019 & \(0.129\pm0.043\) & \(0.083\pm0.031\) & \(0.086\pm0.027\) \\
    \bottomrule
  \end{tabular}
  \vspace{1mm}
 
  \label{MSE_foveation_quantitative}
  \vspace{-5mm}
\end{table}

Fig.~\ref{Evaluate_foveation_main} visualises the foveation maps from different variants of our method on examples from two datasets. In general, `Ours-Mean' and `Ours-GSM' predicted foveation maps visually similar to the corresponding 'Gold Standard', illustrating that the foveation module has generally learned to provide higher resolution patches where needed. This aligns well with our motivation as explained in Fig.~\ref{fig:problem_highlight}, that single fixed path size is not optimal across different spatial locations. We also report the mean-squared-error (MSE) between the learnt foveation and the gold standard in Table.~\ref{MSE_foveation_quantitative}. In general the results are consistent with the qualitative observations shown in Fig.~\ref{Evaluate_foveation_main}.

\vspace{-1mm}
\section{Conclusion}
\vspace{-2mm}
In this work, we propose a new approach for segmenting ultra-high resolution images. In particular, we introduce \textit{foveation module}, a learnable ``dataloader'' which, for a given image, adaptively provide the downstream segmentation model with input patches of the appropriate configuration (FoV/resolution trade-off) at different locations. Such dataloader is trained end-to-end together with the segmentation model to maximise the task performance. We show our approach can improve performance consistently on three public datasets. Our method is simple to implement, requiring simply the addition of the foveation module to an existing segmentation network. A key limitation of the current approach is that only a discrete set of patches are considered --- future work aims to extend this to the continuous settings by
adapting the learnable deformation ideas such as \cite{recasens2018learning,dalca2019learning} to extremely high-resolution images.

\subsection*{Acknowledgements.}
We acknowledge Hongxiang Lin for the insightful discussions, Marnix Jansen for his clinical advice. We are also grateful for the EPSRC grants EP/R006032/1, EP/M020533/1, the CRUK/EPSRC grant NS/A000069/1, and the NIHR UCLH Biomedical Research Centre which supported this research  \looseness=-1

\bibliographystyle{unsrt}  
\bibliography{reference}

\begin{thebibliography}{10}

\bibitem{cordts2016cityscapes}
Marius Cordts, Mohamed Omran, Sebastian Ramos, Timo Rehfeld, Markus Enzweiler,
  Rodrigo Benenson, Uwe Franke, Stefan Roth, and Bernt Schiele.
\newblock The cityscapes dataset for semantic urban scene understanding.
\newblock In {\em Proceedings of the IEEE conference on computer vision and
  pattern recognition}, pages 3213--3223, 2016.

\bibitem{maggiori2017can}
Emmanuel Maggiori, Yuliya Tarabalka, Guillaume Charpiat, and Pierre Alliez.
\newblock Can semantic labeling methods generalize to any city? the inria
  aerial image labeling benchmark.
\newblock In {\em 2017 IEEE International Geoscience and Remote Sensing
  Symposium (IGARSS)}, pages 3226--3229. IEEE, 2017.

\bibitem{srinidhi2019deep}
Chetan~L Srinidhi, Ozan Ciga, and Anne~L Martel.
\newblock Deep neural network models for computational histopathology: A
  survey.
\newblock {\em arXiv preprint arXiv:1912.12378}, 2019.

\bibitem{zhao2017pyramid}
Hengshuang Zhao, Jianping Shi, Xiaojuan Qi, Xiaogang Wang, and Jiaya Jia.
\newblock Pyramid scene parsing network.
\newblock In {\em Proceedings of the IEEE conference on computer vision and
  pattern recognition}, pages 2881--2890, 2017.

\bibitem{zhao2017self}
Jian Zhao, Jianshu Li, Xuecheng Nie, Fang Zhao, Yunpeng Chen, Zhecan Wang,
  Jiashi Feng, and Shuicheng Yan.
\newblock Self-supervised neural aggregation networks for human parsing.
\newblock In {\em Proceedings of the IEEE Conference on Computer Vision and
  Pattern Recognition Workshops}, pages 7--15, 2017.

\bibitem{seth2019automated}
Nikhil Seth, Shazia Akbar, Sharon Nofech-Mozes, Sherine Salama, and Anne~L
  Martel.
\newblock Automated segmentation of dcis in whole slide images.
\newblock In {\em European Congress on Digital Pathology}, pages 67--74.
  Springer, 2019.

\bibitem{chen2016attention}
Liang-Chieh Chen, Yi~Yang, Jiang Wang, Wei Xu, and Alan~L Yuille.
\newblock Attention to scale: Scale-aware semantic image segmentation.
\newblock In {\em Proceedings of the IEEE conference on computer vision and
  pattern recognition}, pages 3640--3649, 2016.

\bibitem{chen2019collaborative}
Wuyang Chen, Ziyu Jiang, Zhangyang Wang, Kexin Cui, and Xiaoning Qian.
\newblock Collaborative global-local networks for memory-efficient segmentation
  of ultra-high resolution images.
\newblock In {\em Proceedings of the IEEE Conference on Computer Vision and
  Pattern Recognition}, pages 8924--8933, 2019.

\bibitem{kamnitsas2017efficient}
Konstantinos Kamnitsas, Christian Ledig, Virginia~FJ Newcombe, Joanna~P
  Simpson, Andrew~D Kane, David~K Menon, Daniel Rueckert, and Ben Glocker.
\newblock Efficient multi-scale 3d cnn with fully connected crf for accurate
  brain lesion segmentation.
\newblock {\em Medical image analysis}, 36:61--78, 2017.

\bibitem{li2019classification}
Yuqian Li, Junmin Wu, and Qisong Wu.
\newblock Classification of breast cancer histology images using multi-size and
  discriminative patches based on deep learning.
\newblock {\em IEEE Access}, 7:21400--21408, 2019.

\bibitem{chen2014semantic}
Liang-Chieh Chen, George Papandreou, Iasonas Kokkinos, Kevin Murphy, and Alan~L
  Yuille.
\newblock Semantic image segmentation with deep convolutional nets and fully
  connected crfs.
\newblock {\em arXiv preprint arXiv:1412.7062}, 2014.

\bibitem{hariharan2015hypercolumns}
Bharath Hariharan, Pablo Arbel{\'a}ez, Ross Girshick, and Jitendra Malik.
\newblock Hypercolumns for object segmentation and fine-grained localization.
\newblock In {\em Proceedings of the IEEE conference on computer vision and
  pattern recognition}, pages 447--456, 2015.

\bibitem{lin2017refinenet}
Guosheng Lin, Anton Milan, Chunhua Shen, and Ian Reid.
\newblock Refinenet: Multi-path refinement networks for high-resolution
  semantic segmentation.
\newblock In {\em Proceedings of the IEEE conference on computer vision and
  pattern recognition}, pages 1925--1934, 2017.

\bibitem{ronneberger2015u}
Olaf Ronneberger, Philipp Fischer, and Thomas Brox.
\newblock U-net: Convolutional networks for biomedical image segmentation.
\newblock In {\em International Conference on Medical image computing and
  computer-assisted intervention}, pages 234--241. Springer, 2015.

\bibitem{noh2015learning}
Hyeonwoo Noh, Seunghoon Hong, and Bohyung Han.
\newblock Learning deconvolution network for semantic segmentation.
\newblock In {\em Proceedings of the IEEE international conference on computer
  vision}, pages 1520--1528, 2015.

\bibitem{badrinarayanan2017segnet}
Vijay Badrinarayanan, Alex Kendall, and Roberto Cipolla.
\newblock Segnet: A deep convolutional encoder-decoder architecture for image
  segmentation.
\newblock {\em IEEE transactions on pattern analysis and machine intelligence},
  39(12):2481--2495, 2017.

\bibitem{lin2017feature}
Tsung-Yi Lin, Piotr Doll{\'a}r, Ross Girshick, Kaiming He, Bharath Hariharan,
  and Serge Belongie.
\newblock Feature pyramid networks for object detection.
\newblock In {\em Proceedings of the IEEE conference on computer vision and
  pattern recognition}, pages 2117--2125, 2017.

\bibitem{sun2019high}
Ke~Sun, Yang Zhao, Borui Jiang, Tianheng Cheng, Bin Xiao, Dong Liu, Yadong Mu,
  Xinggang Wang, Wenyu Liu, and Jingdong Wang.
\newblock High-resolution representations for labeling pixels and regions.
\newblock {\em arXiv preprint arXiv:1904.04514}, 2019.

\bibitem{xiao2018unified}
Tete Xiao, Yingcheng Liu, Bolei Zhou, Yuning Jiang, and Jian Sun.
\newblock Unified perceptual parsing for scene understanding.
\newblock In {\em Proceedings of the European Conference on Computer Vision
  (ECCV)}, pages 418--434, 2018.

\bibitem{girshick2014rich}
Ross Girshick, Jeff Donahue, Trevor Darrell, and Jitendra Malik.
\newblock Rich feature hierarchies for accurate object detection and semantic
  segmentation.
\newblock In {\em Proceedings of the IEEE conference on computer vision and
  pattern recognition}, pages 580--587, 2014.

\bibitem{girshick2015fast}
Ross Girshick.
\newblock Fast r-cnn.
\newblock In {\em Proceedings of the IEEE international conference on computer
  vision}, pages 1440--1448, 2015.

\bibitem{he2017mask}
Kaiming He, Georgia Gkioxari, Piotr Doll{\'a}r, and Ross Girshick.
\newblock Mask r-cnn.
\newblock In {\em Proceedings of the IEEE international conference on computer
  vision}, pages 2961--2969, 2017.

\bibitem{recasens2018learning}
Adria Recasens, Petr Kellnhofer, Simon Stent, Wojciech Matusik, and Antonio
  Torralba.
\newblock Learning to zoom: a saliency-based sampling layer for neural
  networks.
\newblock In {\em Proceedings of the European Conference on Computer Vision
  (ECCV)}, pages 51--66, 2018.

\bibitem{girdhar2016learning}
Rohit Girdhar, David~F Fouhey, Mikel Rodriguez, and Abhinav Gupta.
\newblock Learning a predictable and generative vector representation for
  objects.
\newblock In {\em European Conference on Computer Vision}, pages 484--499.
  Springer, 2016.

\bibitem{choy20163d}
Christopher~B Choy, Danfei Xu, JunYoung Gwak, Kevin Chen, and Silvio Savarese.
\newblock 3d-r2n2: A unified approach for single and multi-view 3d object
  reconstruction.
\newblock In {\em European conference on computer vision}, pages 628--644.
  Springer, 2016.

\bibitem{wang2018pixel2mesh}
Nanyang Wang, Yinda Zhang, Zhuwen Li, Yanwei Fu, Wei Liu, and Yu-Gang Jiang.
\newblock Pixel2mesh: Generating 3d mesh models from single rgb images.
\newblock In {\em Proceedings of the European Conference on Computer Vision
  (ECCV)}, pages 52--67, 2018.

\bibitem{gkioxari2019mesh}
Georgia Gkioxari, Jitendra Malik, and Justin Johnson.
\newblock Mesh r-cnn.
\newblock In {\em Proceedings of the IEEE International Conference on Computer
  Vision}, pages 9785--9795, 2019.

\bibitem{mescheder2019occupancy}
Lars Mescheder, Michael Oechsle, Michael Niemeyer, Sebastian Nowozin, and
  Andreas Geiger.
\newblock Occupancy networks: Learning 3d reconstruction in function space.
\newblock In {\em Proceedings of the IEEE Conference on Computer Vision and
  Pattern Recognition}, pages 4460--4470, 2019.

\bibitem{tatarchenko2017octree}
Maxim Tatarchenko, Alexey Dosovitskiy, and Thomas Brox.
\newblock Octree generating networks: Efficient convolutional architectures for
  high-resolution 3d outputs.
\newblock In {\em Proceedings of the IEEE International Conference on Computer
  Vision}, pages 2088--2096, 2017.

\bibitem{marin2019efficient}
Dmitrii Marin, Zijian He, Peter Vajda, Priyam Chatterjee, Sam Tsai, Fei Yang,
  and Yuri Boykov.
\newblock Efficient segmentation: Learning downsampling near semantic
  boundaries.
\newblock In {\em Proceedings of the IEEE International Conference on Computer
  Vision}, pages 2131--2141, 2019.

\bibitem{katharopoulos2019processing}
Angelos Katharopoulos and Fran{\c{c}}ois Fleuret.
\newblock Processing megapixel images with deep attention-sampling models.
\newblock {\em arXiv preprint arXiv:1905.03711}, 2019.

\bibitem{shen2019globally}
Yiqiu Shen, Nan Wu, Jason Phang, Jungkyu Park, Gene Kim, Linda Moy, Kyunghyun
  Cho, and Krzysztof~J Geras.
\newblock Globally-aware multiple instance classifier for breast cancer
  screening.
\newblock In {\em International Workshop on Machine Learning in Medical
  Imaging}, pages 18--26. Springer, 2019.

\bibitem{DeepGlobe18}
Ilke Demir, Krzysztof Koperski, David Lindenbaum, Guan Pang, Jing Huang, Saikat
  Basu, Forest Hughes, Devis Tuia, and Ramesh Raskar.
\newblock Deepglobe 2018: A challenge to parse the earth through satellite
  images.
\newblock In {\em The IEEE Conference on Computer Vision and Pattern
  Recognition (CVPR) Workshops}, June 2018.

\bibitem{Gleason2019}
Gleason 2019 challenge.
\newblock \url{https://gleason2019.grand-challenge.org/Home/}.
\newblock Accessed: 2020-02-30.

\bibitem{maddison2016concrete}
Chris~J Maddison, Andriy Mnih, and Yee~Whye Teh.
\newblock The concrete distribution: A continuous relaxation of discrete random
  variables.
\newblock {\em arXiv preprint arXiv:1611.00712}, 2016.

\bibitem{jang2016categorical}
Eric Jang, Shixiang Gu, and Ben Poole.
\newblock Categorical reparameterization with gumbel-softmax.
\newblock {\em arXiv preprint arXiv:1611.01144}, 2016.

\bibitem{gal2017concrete}
Yarin Gal, Jiri Hron, and Alex Kendall.
\newblock Concrete dropout.
\newblock In {\em Advances in neural information processing systems}, pages
  3581--3590, 2017.

\bibitem{bragman2019stochastic}
Felix~JS Bragman, Ryutaro Tanno, Sebastien Ourselin, Daniel~C Alexander, and
  Jorge Cardoso.
\newblock Stochastic filter groups for multi-task cnns: Learning specialist and
  generalist convolution kernels.
\newblock In {\em Proceedings of the IEEE International Conference on Computer
  Vision}, pages 1385--1394, 2019.

\bibitem{williams1992simple}
Ronald~J Williams.
\newblock Simple statistical gradient-following algorithms for connectionist
  reinforcement learning.
\newblock {\em Machine learning}, 8(3-4):229--256, 1992.

\bibitem{tucker2017rebar}
George Tucker, Andriy Mnih, Chris~J Maddison, John Lawson, and Jascha
  Sohl-Dickstein.
\newblock Rebar: Low-variance, unbiased gradient estimates for discrete latent
  variable models.
\newblock In {\em Advances in Neural Information Processing Systems}, pages
  2627--2636, 2017.

\bibitem{grathwohl2017backpropagation}
Will Grathwohl, Dami Choi, Yuhuai Wu, Geoffrey Roeder, and David Duvenaud.
\newblock Backpropagation through the void: Optimizing control variates for
  black-box gradient estimation.
\newblock {\em arXiv preprint arXiv:1711.00123}, 2017.

\bibitem{ilse2018attention}
Maximilian Ilse, Jakub~M Tomczak, and Max Welling.
\newblock Attention-based deep multiple instance learning.
\newblock {\em arXiv preprint arXiv:1802.04712}, 2018.

\bibitem{xu2015show}
Kelvin Xu, Jimmy Ba, Ryan Kiros, Kyunghyun Cho, Aaron Courville, Ruslan
  Salakhudinov, Rich Zemel, and Yoshua Bengio.
\newblock Show, attend and tell: Neural image caption generation with visual
  attention.
\newblock In {\em International conference on machine learning}, pages
  2048--2057, 2015.

\bibitem{bengio2013estimating}
Yoshua Bengio, Nicholas L{\'e}onard, and Aaron Courville.
\newblock Estimating or propagating gradients through stochastic neurons for
  conditional computation.
\newblock {\em arXiv preprint arXiv:1308.3432}, 2013.

\bibitem{van2017neural}
Aaron van~den Oord, Oriol Vinyals, et~al.
\newblock Neural discrete representation learning.
\newblock In {\em Advances in Neural Information Processing Systems}, pages
  6306--6315, 2017.

\bibitem{dalca2019learning}
Adrian Dalca, Marianne Rakic, John Guttag, and Mert Sabuncu.
\newblock Learning conditional deformable templates with convolutional
  networks.
\newblock In {\em Advances in neural information processing systems}, pages
  806--818, 2019.

\bibitem{warfield2004simultaneous}
Simon~K Warfield, Kelly~H Zou, and William~M Wells.
\newblock Simultaneous truth and performance level estimation (staple): an
  algorithm for the validation of image segmentation.
\newblock {\em IEEE transactions on medical imaging}, 23(7):903--921, 2004.

\bibitem{he2015delving}
Kaiming He, Xiangyu Zhang, Shaoqing Ren, and Jian Sun.
\newblock Delving deep into rectifiers: Surpassing human-level performance on
  imagenet classification.
\newblock In {\em Proceedings of the IEEE international conference on computer
  vision}, pages 1026--1034, 2015.

\bibitem{kingma2014adam}
Diederik~P Kingma and Jimmy Ba.
\newblock Adam: A method for stochastic optimization.
\newblock {\em arXiv preprint arXiv:1412.6980}, 2014.

\bibitem{liu2015parsenet}
Wei Liu, Andrew Rabinovich, and Alexander~C Berg.
\newblock Parsenet: Looking wider to see better.
\newblock {\em arXiv preprint arXiv:1506.04579}, 2015.

\end{thebibliography}


\appendix
\counterwithin{figure}{subsection}
\counterwithin{table}{subsection}
\newpage
\begin{center}
\LARGE{\textbf{Supplementary Material: \\Foveation for Segmentaiton of \\Ultra-high Resolution Images}}
\end{center}

\label{section_supplementary}


\section{Details of Datasets, Architectures, Training and Baselines}
\label{method}

\subsection{Datasets}
\label{dataset}
In this work, we verified our method on three segmentation datasets: DeepGlobe aerial scenes segmentation dataset \cite{DeepGlobe18}, CityScape urban scenes segmentation dataset \cite{cordts2016cityscapes} and Gleason2019 medical histopathological segmentation dataset \cite{Gleason2019}.

The \textbf{DeepGlobe \cite{DeepGlobe18}} dataset has 803 high-resolution ($2448\times2448$ pixels) images of aerial scenes. There are 7 classes of dense annotations, 6 classes among them are used for training and evaluation according to \cite{DeepGlobe18}. We randomly split the dataset into train, validate and test with 455, 207, and 142 images respectively.

The \textbf{CityScape \cite{cordts2016cityscapes}} dataset contains 5000 high-resolution ($2048\times1024$ pixels) urban scenes images collected across 27 European Cities. The finely-annotated images contain 30 classes, and 19 classes among them are used for training and evaluation according to \cite{cordts2016cityscapes}. The 5000 images from the CityScape are divided into 2975/500/1525 images for training, validation and testing. 

The \textbf{Gleason2019 \cite{Gleason2019}} dataset contains 322 high-resolution ($5000\times5000$ pixels) medical histopathological images. Each image is finely-annotated by a subset of 6 annotators (all experts), which annotates each pixel into one from four classes (Benign, Gleason Grade 3,4,5). Preprocessing for emprical analysis: A subset of 298 training examples have been used in the first part of our work. We fuse 6 annotations into 1 using pixel-level probabilistic analysis by STAPLE \cite{warfield2004simultaneous}. Each image is paired with 1 STAPLE fused annotation as gold standard (to be used as ground truth during training and evaluation). Pre-processing for foveation experiments: on top of 298 pre-processed images in empirical analysis, we extract the central part of input images of size $2448\times2448$ pixels from the original histology images to ensure input images have constant size.

\subsection{Network Architectures and Implementation Details}
\label{Architectures_and_Implementation}

\textbf{Architectures:} The \textit{foveation module} is defined as a small CNN architecture comprised of 3 convolution layers, each with $3\times3$ kernels follower by BatchNorm and Relu. The number of kernels in each respective layer is $\{40,40,5\}$. A softmax layer is added at the end. All convolution layers are initialised following He initialization \cite{he2015delving}. The Segmentation module was defined as a deep CNN architecture , with HRNetV2-W48 \cite{sun2019high} applied in Cityscape and Gleason2019 dataset, and UPerNet \cite{xiao2018unified} applied in DeepGlobe dataset (details provided in the original literature). The segmentation network HRNetV2-W48 is pre-trained on Imagenet dataset as provided by the author \cite{sun2019high}. 
\vspace{-1mm}

\textbf{Foveation module specific:} In all settings, Patch-Extractor $PE(\mathbf{I},i)$ extracts, at each location $i$, a set of 5 patches at different FoVs of $\{500^2, 1000^2, 1500^2, 2000^2, 2500^2\}$ for DeepGlobe, $\{512^2, 640^2, 768^2, 896^2, 1024^2\}$ for Cityscape and $\{1100^2, 2000^2, 2900^2, 3800^2, 4400^2\}$ for Gleason2019), all patches in each dataset are downsampled to the same size as smallest patch in the original resolution. When generating low-resolution counterparts $\mathbf{I}_{\text{lr}}$, downsampling rates of 1/24, 1/16, 1/44 are applied for DeepGlobe, Cityscape and Gleason2019 respectively, for both training and inference, unless otherwise stated.

\textbf{Minibatch construction:}
In our experiments, each minibatch of size $P$ consists of a set of $B$ patches extracted from each of $L$ different mega-pixel images (i.e., $P = L*B)$. More specifically, for each mega-pixel image $\mathbf{I}$, we select a subset of $B$ locations from its low resolution counterpart $\mathbf{I}_{\text{lr}}$ and extract patches from them. Therefore for a set of $L$ mega-pixel images, a total number of $P = L*B$ patches are sampled and passed to the segmentation network.  We keep the total number of patches, $P$ fixed in each dataset ($P=8$ for DeepGlobe and Cityscape, and $P=2$ Gleason2019) to max out the available GPU resources, while we selected the combination of $B$ and $L$ according to the validation performance for each model instance.

\subsection{Training}
\label{Training}
For all experiments, we employ the same training scheme unless otherwise stated. We optimize parameters using Adam \cite{kingma2014adam} with initial learning rate of $2e10^{-5}$ and $\beta=0.9$, and train for 50 epochs on DeepGlobe and Gleason2019 dataset, and 100 epoches on Cityscape dataset. Following prior works \cite{liu2015parsenet}, we adopt ‘poly’ learning rate policy and the power is set to 0.9. We set maximum iteration number to 410K for experiments on the DeepGlobe dataset, 595K for Cityscapes and 95K for Gleason2019. Segmentation networks are trained on 2 or 4 GPUs with syncBN. The temperature term  $\tau$ in the Gumbel-Softmax gradient estimator is annealed by scheduler $\tau = \text{max}(0.10, \exp(-rt))$ as recommended in \cite{jang2016categorical}, where $r$ is the annealing rate and $t$ is the current training iteration. We used $r=1/$\textit{'total iteration number'} for our models.

\subsection{Baselines}
\label{result:Baselines}
For each of the three datasets, we compare our method against a variety of baselines. Firstly, we consider the same segmentation networks (HRNet or UPerNet), but trained on input patches of fixed FoV/resolution-tradeoffs. Each case was implemented by setting the outputs of the foveation module $F_{\theta}(\mathbf{x}_{\text{lr}})$ to a fixed one-hot vectors, thus selecting only one scale per baseline from the given set of 5 patches. Additionally, on DeepGlobe and Gleason2019 datasets, to illustrate our foveation approach do better than random guess or uniform average, two additional baselines are also applied: a uniform random one-hot baselines that randomly select one from the given set of 5 patches with varying FoV/resolution, and an average baseline that assign equal probability $f_{d}^{(i)}$ of 1/5 over the set of 5 patches with varying FoV/resolution. Lastly, to examine the robustness of our approach, we also calculate the performance of the ensemble model by averaging the five one-hot baseline predictions.

\section{Additional Results}
\label{result}


\subsection{Patch Configuration Matters in Segmentation}
\label{method:Patch_Configuration_ Matters}

\begin{figure}[H]
\centering
\includegraphics[width=0.7\linewidth]{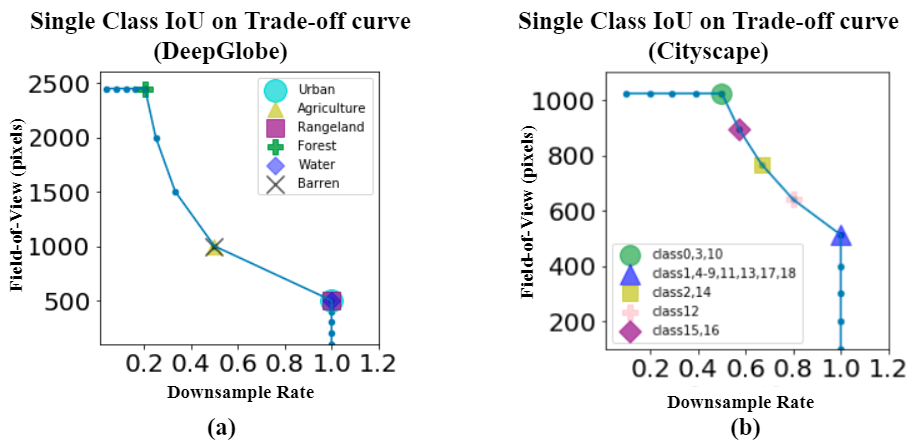}
\caption{Quantitative evidence that the optimal FoV/resolution trade-off varies over classes on (a) DeepGlobe dataset and (b) Cityscape dataset: segmentation networks trained with different trade-off configurations where best performance for each class is highlighted (see Legends for references). Cityscape class code: 0-road, 1-sidewalk, 2-building, 3-wall, 4-fence, 5-pole, 6-traffic light, 7-raffic sign, 8-vegetation, 9-terrain, 10-sky, 11-person, 12-rider, 13-car, 14-truck, 15-bus, 16-train, 17-motorcycle, 18-bicycle. } \label{fig:Deepglobe_Cityscape_average_patch_size_against_optimal} 
\end{figure}

To further illustrate the impact of the patch FoV/resolution trade-off on the segmentation performance and its spatial variation across the image.  Fig.~\ref{fig:Deepglobe_Cityscape_average_patch_size_against_optimal} shows the results of the same motivational experiment as in the main text Fig. 2(a), but on the DeepGlobe and Cityscape datasets. For each data set, we first train a set of different segmentation networks, each network with a different combination of FoV and downsampling rate (i.e., resolution); see the small blue dots on the curve in Fig.~\ref{fig:Deepglobe_Cityscape_average_patch_size_against_optimal}. We note that the maximum tensor size of the input patch is capped at $500^2$ for DeepGlobe and $512^2$ for Cityscape, and constant along the curve in Fig.~\ref{fig:Deepglobe_Cityscape_average_patch_size_against_optimal}. The segmentation network with the best performance for each class is highlighted in different shaped marks (see Fig.~\ref{fig:Deepglobe_Cityscape_average_patch_size_against_optimal}). It shows again that there is no ``one-size-fits-all'' patch configuration of the training data that leads to the best performance overall and for all individual classes. 

\subsection{Additional Quantitative Comparison on Cityscape and Gleason 2019}
\label{result:performance_comparison}

The class-wise segmentation performance on Cityscape are given in Table~\ref{results: Performance_cityscape_1} and Table~\ref{results: Performance_cityscape_2}. These results show that our approach improved fixed patch size baselines by large margin for both overall average mIoU (+6\%) and most single class IoU.

\begin{table}[H]
  \centering
  \vspace{-3mm}
  \scriptsize
  
  \begin{tabular}{lllllllllll}
    \toprule
    Dataset & \multicolumn{10}{c}{Cityscape}  \\
    \midrule 
    {Class} & All & road & sidewalk & building & wall & fence & pole & {traffic light} & raffic sign & vegetation \\
    \midrule
    Baseline-Patch 1 & 67.7 & 97.6 & 80.8 & 89.8 & 49.7 & 51.4 & 53.9 & 39.0 & 63.3 & 89.6 \\
	Baseline-Patch 2 & 68.0 & 97.3 & 81.0 & 89.5 & 41.9 & 52.2 & 51.5 & 32.9 & 66.0 & 89.2 \\
	Baseline-Patch 3 & 69.8 & 97.5 & 81.2 & 90.3 & 47.0 & 55.8 & 54.2 & 46.8 & 68.6 & 90.0 \\
	Baseline-Patch 4 & 70.6 & 97.0 & 82.1 & 89.8 & 47.3 & 53.4 & 57.6 & 56.9 & 72.1 & 90.3 \\
	Baseline-Patch 5 & 70.2 & 95.0 & \cellcolor{blue!15}84.4 & 90.3 & 45.6 & 56.2 & 63.4 & 61.4 & \cellcolor{blue!15}76.5 & 91.4 \\
	Baseline-Random & 50.3 & 94.1 & 68.1 & 80.7 & 36.5 & 40.2 & 18.8 & 13.3 & 31.8 & 79.6 \\
	Baseline-Average & 56.6 & 96.3 & 73.6 & 85.7 & 31.6 & 40.4 & 39.0 & 19.8 & 49.8 & 86.5 \\
	Ensemble & \cellcolor{blue!15}72.5 & \cellcolor{blue!15}97.7 & 83.8 & \cellcolor{blue!15}91.1 & \cellcolor{blue!15}51.7 & \cellcolor{blue!15}58.2 & 58.9 & 50.9 & 72.6 & 90.9 \\
	\midrule
	Ours-Mean & \cellcolor{red!15} {\bfseries76.1} & \cellcolor{red!15} {\bfseries98.2} & \cellcolor{red!15} {\bfseries84.9} & \cellcolor{red!15} {\bfseries92.3} & \cellcolor{red!15} {\bfseries55.2} & \cellcolor{red!15} {\bfseries61.4} & \cellcolor{red!15} {\bfseries64.5} & \cellcolor{red!15} {\bfseries71.4} & \cellcolor{red!15} {\bfseries80.1} & \cellcolor{red!15} {\bfseries92.5} \\
	Ours-Mode & 71.4 & 97.5 & 80.8 & 90.8 & 45.4 & 55.3 & 55.6 & 56.5 & 67.7 & 90.7 \\
	Ours-GSM & 70.5 & \cellcolor{blue!15}97.7 & 82.2 & 89.1 & 44.8 & 55.4 & \cellcolor{blue!15}64.5 & \cellcolor{blue!15}66.1 & 76.2 & \cellcolor{blue!15}91.8 \\
    \bottomrule
  \end{tabular} 
\vspace{1mm}
    \caption{\footnotesize Segmentation performance measured in mIoU (\%) for all 19 classes and IoU(\%) for the first 9 classes on Cityscape datasets. The best and the second best results are shown in red and blue.} 

  \vspace{-1mm} 
  \label{results: Performance_cityscape_1}
\end{table}

\begin{table}[H]
  \centering
  \vspace{-3mm}
  \scriptsize
  
  \begin{tabular}{lllllllllll}
    \toprule
    Dataset & \multicolumn{10}{c}{Cityscape}  \\
    \midrule 
    {Class} & terrain & sky & person & rider & car & truck & bus & train & motorcycle & bicycle \\
    \midrule
    Baseline-Patch 1 & 59.4 & 90.4 & 73.9 & 51.6 & 91.9 & 54.9 & 74.0 & 61.3 & 45.3 & 69.1 \\
	Baseline-Patch 2 & 55.9 & 90.1 & 75.3 & 55.4 & 92.0 & 54.1 & \cellcolor{blue!15}80.2 & \cellcolor{red!15} {\bfseries74.8} & 43.7 & 69.8 \\
	Baseline-Patch 3 & 57.4 & 88.8 & 77.1 & 56.9 & 92.8 & 57.4 & 76.8 & 63.6 & 51.5 & 72.5 \\
	Baseline-Patch 4 & 56.7 & 85.6 & 79.0 & 60.0 & 92.8 & 54.0 & 78.2 & 58.1 & 56.8 & 74.2 \\
	Baseline-Patch 5 & 59.9 & 68.9 & \cellcolor{blue!15}81.2 & 58.0 & \cellcolor{blue!15}93.2 & 53.0 & 69.6 & 50.6 & \cellcolor{red!15} {\bfseries59.3} & \cellcolor{blue!15}76.8 \\
	Baseline-Random & 50.7 & 58.4 & 53.8 & 34.4 & 79.1 & 49.3 & 46.3 & 37.8 & 31.2 & 51.6 \\
	Baseline-Average & 47.5 & 89.3 & 65.7 & 43.6 & 89.0 & 47.6 & 45.7 & 38.6 & 23.2 & 61.6 \\
	Ensemble & 60.9 & 90.9 & 79.4 & 60.1 & \cellcolor{blue!15}93.2 & 57.1 & 79.6 & 69.0 & 56.4 & 74.8 \\
	\midrule
	Ours-Mean & \cellcolor{red!15} {\bfseries64.2} & \cellcolor{red!15} {\bfseries94.1} & \cellcolor{red!15} {\bfseries82.8} & \cellcolor{red!15} {\bfseries64.0} & \cellcolor{red!15} {\bfseries94.5} & \cellcolor{blue!15}63.6 & \cellcolor{red!15} {\bfseries81.0} & 63.3 & \cellcolor{blue!15}59.2 & \cellcolor{red!15} {\bfseries78.2} \\
	Ours-Mode & 60.8 & \cellcolor{blue!15}92.1 & 75.5 & 55.8 & 93.1 & \cellcolor{red!15} {\bfseries66.0} & 76.4 & \cellcolor{blue!15}69.9 & 54.5 & 71.6 \\
	Ours-GSM & \cellcolor{blue!15}62.1 & 75.1 & 80.8 & \cellcolor{blue!15}60.8 & 92.8 & 53.4 & 74.8 & 42.6 & 52.9 & 76.6 \\
    \bottomrule
  \end{tabular} 
\vspace{1mm}
    \caption{\footnotesize Segmentation performance measured in mIoU (\%) for all 19 classes and IoU(\%) for the last 10 classes on Cityscape datasets. The best and the second best results are shown in red and blue.} 

  \vspace{-1mm} 
  \label{results: Performance_cityscape_2}
\end{table}

As declared in Section 4.2, on Gleason2019 data set, we compare with published results as follows: we compare ours approaches with the top 2 results on the Gleason2019 challenge leaderboard (\url{https://gleason2019.grand-challenge.org/Results/}) ranked by overall all classes average segmentation accuracy. We also collect highest segmentation accuracy of each classes as a third Single-Class-Best case for comparison. We quantify segmentation performance via pixel accuracy for each class, to be consistent and comparable against results released on the leaderboard. It worth to note that for all results, we remove Gleason Grade 5 in evaluation it is under represented - only  2\% pixels in the given dataset. The results are shown in Table~\ref{gleason_leader_board}, where our model achieves better segmentation accuracy against the top performers in the challenge for the two most clinically important and ambigous classes (Gleason Grade 3 and 4) by 13.1\% and 15.7\%.

\begin{table}[H]
  \caption{Gleason2019: Quantitative comparison to challenge leaderboard results measured by single class pixel accuracy. Best performance of each class is highlighted.}
  \label{gleason_leader_board}
  \centering
  \begin{tabular}{llll}
    \toprule
    Experiment & Benign & Grade 3 & Grade 4 \\
    \midrule
    Our-Mean Approximation & 88.3 & {\bfseries83.8} & 78.1 \\
    Our-GSM & 87.6 & 78.9 & 81.3 \\
    Our-Mode Approximation & 85.2 & 71.5 & {\bfseries86.3} \\
    \midrule
    Overall Top1 & {\bfseries95.9} & 2.24 & 16.5 \\
    Overall Top2 & 83.0 & 52.7 & 54.0 \\
    Single-Class-Best & 95.9 & 70.7 & 70.6 \\
    \bottomrule
  \end{tabular}
\end{table}

\subsection{Additional Qualitative Comparison}
\label{result:visual_comparison}
\begin{figure}[h]
\centering
\includegraphics[width=\textwidth]{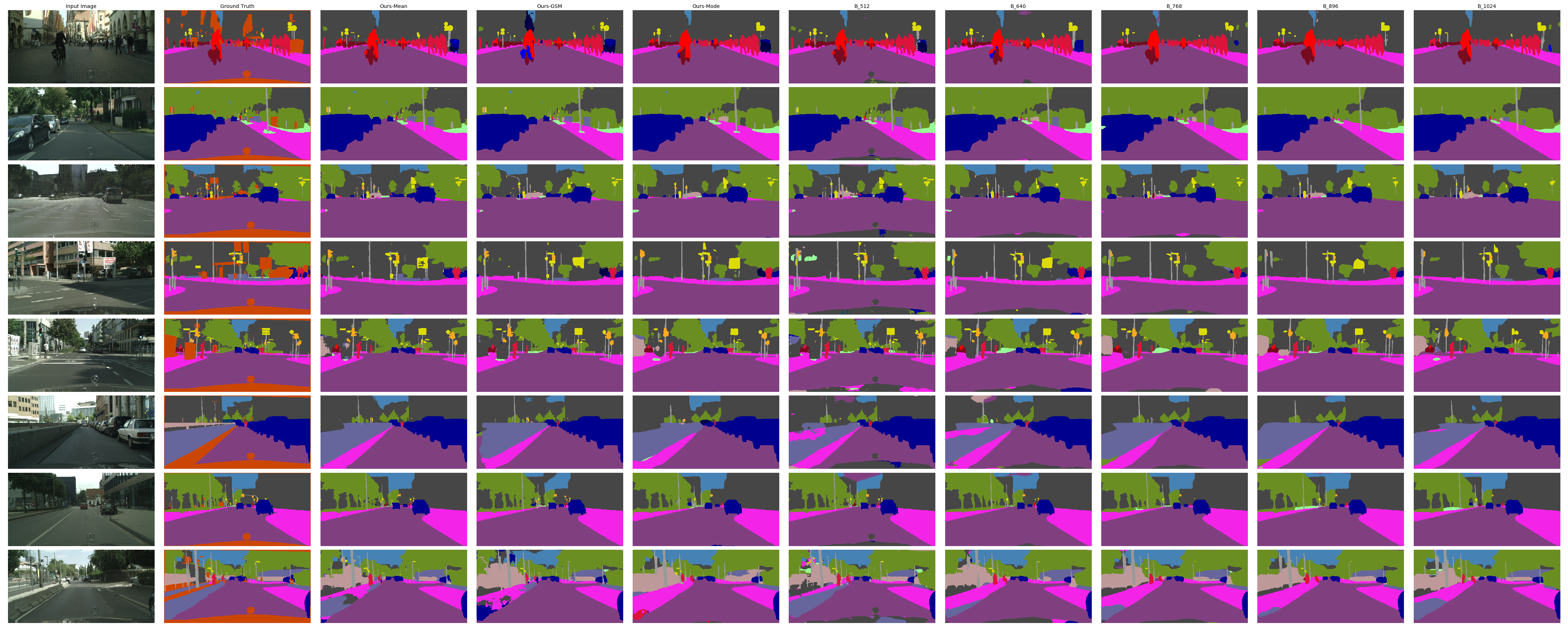}
\caption{More segmentation examples on Cityscape} \label{Cityscape_segmentation_FULL}
\end{figure}

\begin{figure}[h]
\centering
\includegraphics[width=\textwidth]{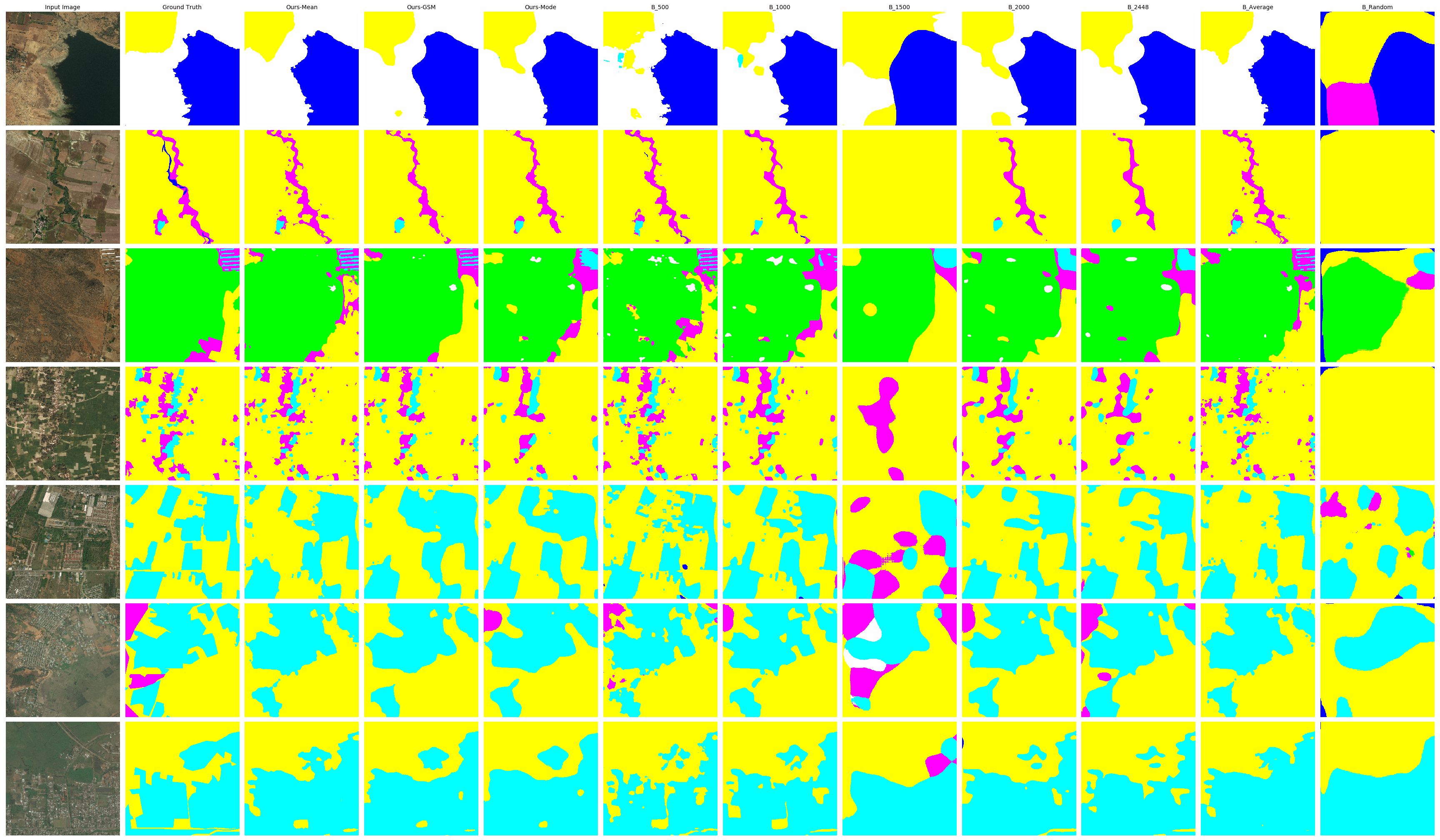}
\caption{More segmentation examples on DeepGlobe} \label{DeepGlobe_segmentation_FULL}
\end{figure}

\begin{figure}[h]
\centering
\includegraphics[width=\textwidth]{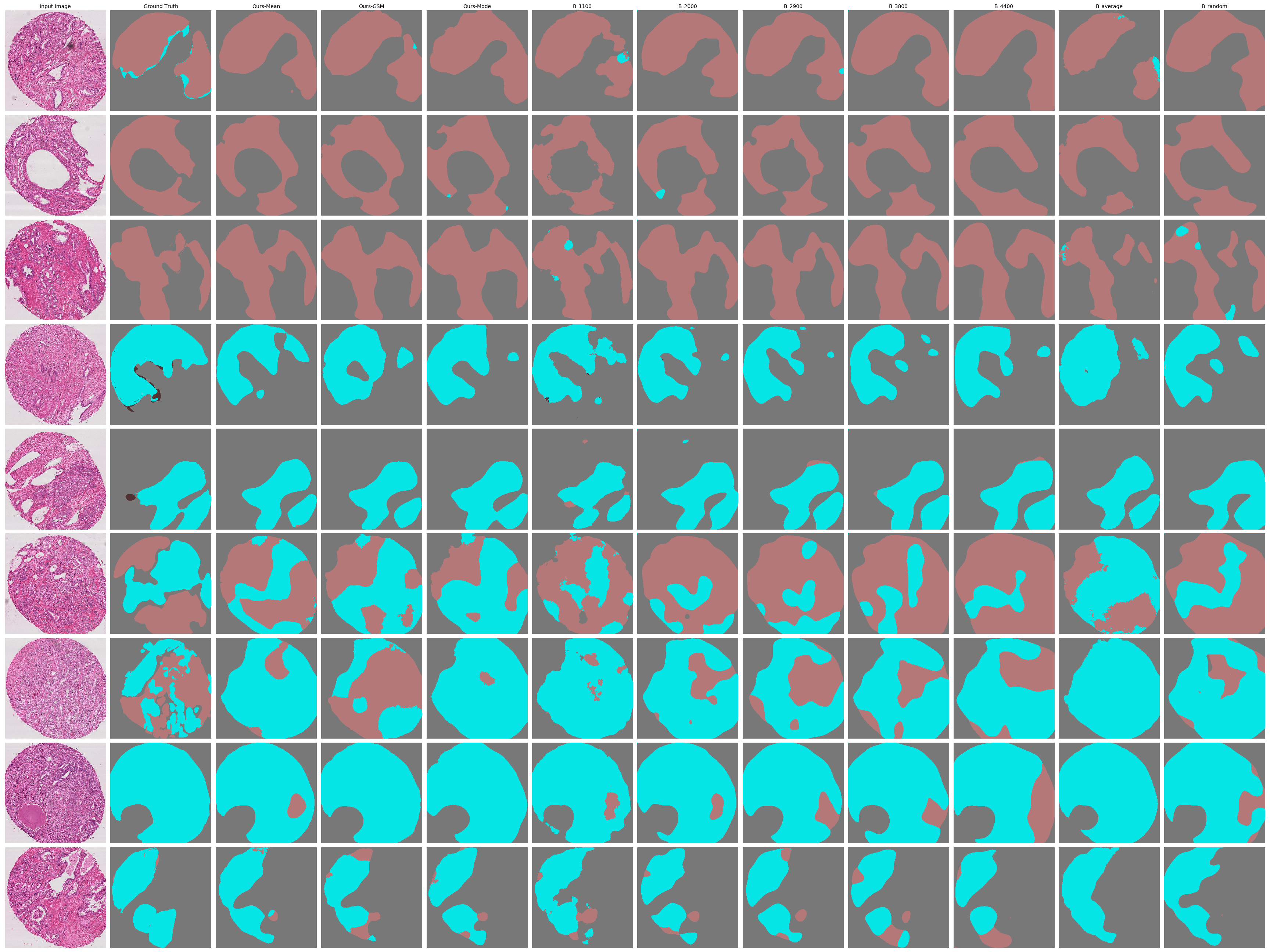}
\caption{More segmentation examples on Gleason2019}  \label{Gleason2019_Qualitative_FULL}
\end{figure}

More qualitative segmentation examples are given in Fig.~\ref{Cityscape_segmentation_FULL} for Cityscape data set, Fig.~\ref{DeepGlobe_segmentation_FULL} for DeepGlobe data set and Fig.~\ref{Gleason2019_Qualitative_FULL} for Gleason2019 data set.

\subsection{Evaluation of Foveation Module}
\label{result:evaluation_foveation}

Here we give details on calculating foveation map and the performance weighted 'Gold standard'. For one input image  $\mathbf{I}$, at each $i^{\text{th}}$ pixel in $\mathbf{I}_{\text{lr}}$, we use the foveation output, a D-dimensional probability vector $f^{(i)}_{\theta}(\textbf{I}_\text{lr}) = [f_{\theta,1}^{(i)}(\textbf{I}_\text{lr}), ..., f_{\theta,D}^{(i)}(\textbf{I}_\text{lr})]$, to weight average the given D FoVs (corresponds to the given set of D patches), referred to as weighted average FoVs. To visualise the spatial distribution of FoV/Resolution Trade-off, we calculate and plot weighted average FoVs over all location in $\mathbf{I}_{\text{lr}}$ and refer it as the "foveation map". For consistency and better visualisation, we apply minimum-maximum normalisation to 0-1 based on the smallest and largest FoV in the given set of D patches, where 1 indicting largest FoV-lowest resolution, and 0 indicting smallest FoV-highest resolution.

We also use the similar strategy to visualise FoVs weighted by performance (mIoU). Specifically, at each $i^{\text{th}}$ pixel in $\mathbf{I}_{\text{lr}}$, we use the segmentation performance (mIoU) ${m}(\mathbf{x}^{(i)}; \phi)$ from the five fixed patch baselines to weight average the FoVs of the given set of patches and plot in the same way as foveation map, which is referred to as 'Gold Standard' to compare with foveation map. It is worth noting that this 'Gold Standard' is not real ground truth as only limited patch size are evaluated. 

\begin{figure}[h]
\centering
\includegraphics[width=0.82\textwidth]{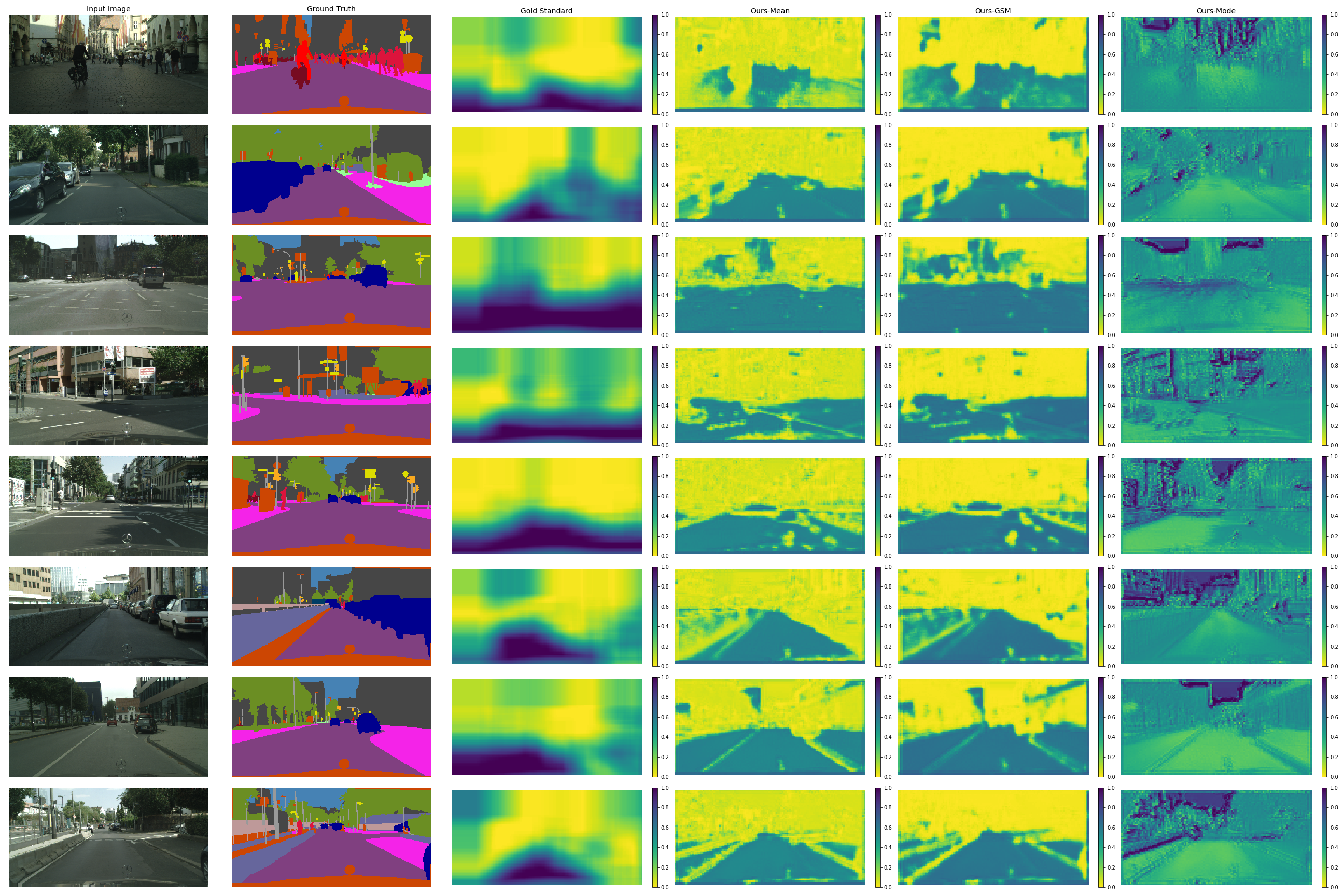}
\caption{More qualitative results of the foveation maps on Cityscape} \label{Cityscape_FoV_Eval_FULL}
\end{figure}

\begin{figure}[H]
\centering
\includegraphics[width=\textwidth]{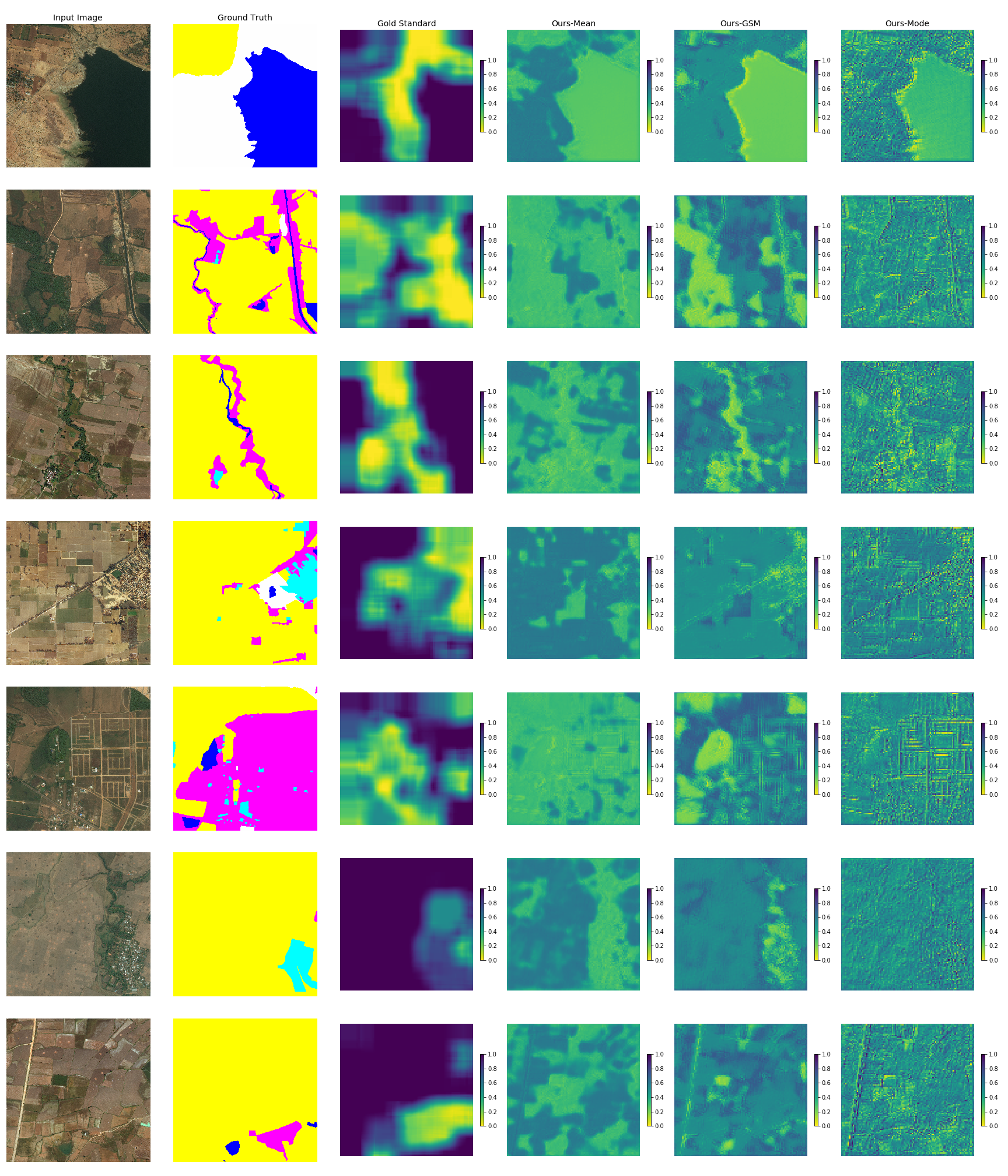}
\caption{More qualitative results of the foveation maps on DeepGlobe} \label{Deepglobe_FoV_Eval_FULL}
\end{figure}

\begin{figure}[H]
\centering
\includegraphics[width=\textwidth]{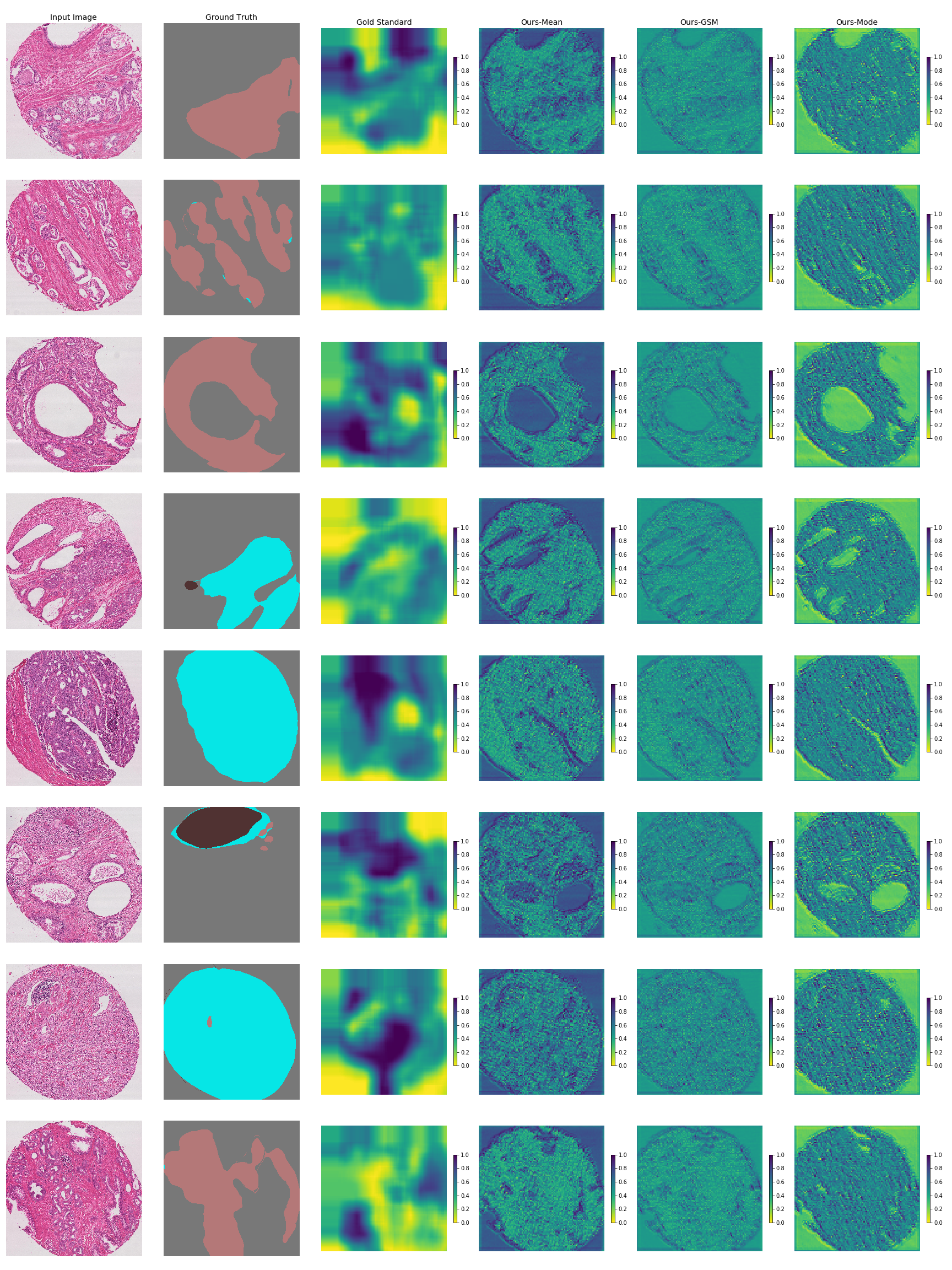}
\caption{More qualitative results of the foveation maps on Gleason2019} \label{Gleason_FoV_Eval_FULL}
\end{figure}

Fig.~\ref{Cityscape_FoV_Eval_FULL}, Fig.~\ref{Deepglobe_FoV_Eval_FULL} and Fig.~\ref{Gleason_FoV_Eval_FULL} gives diverse examples of visual comparison between performance weighted 'Gold Standard' to "foveation maps" generated with ours approaches.

\begin{figure}[H]
\centering
\includegraphics[width=\linewidth]{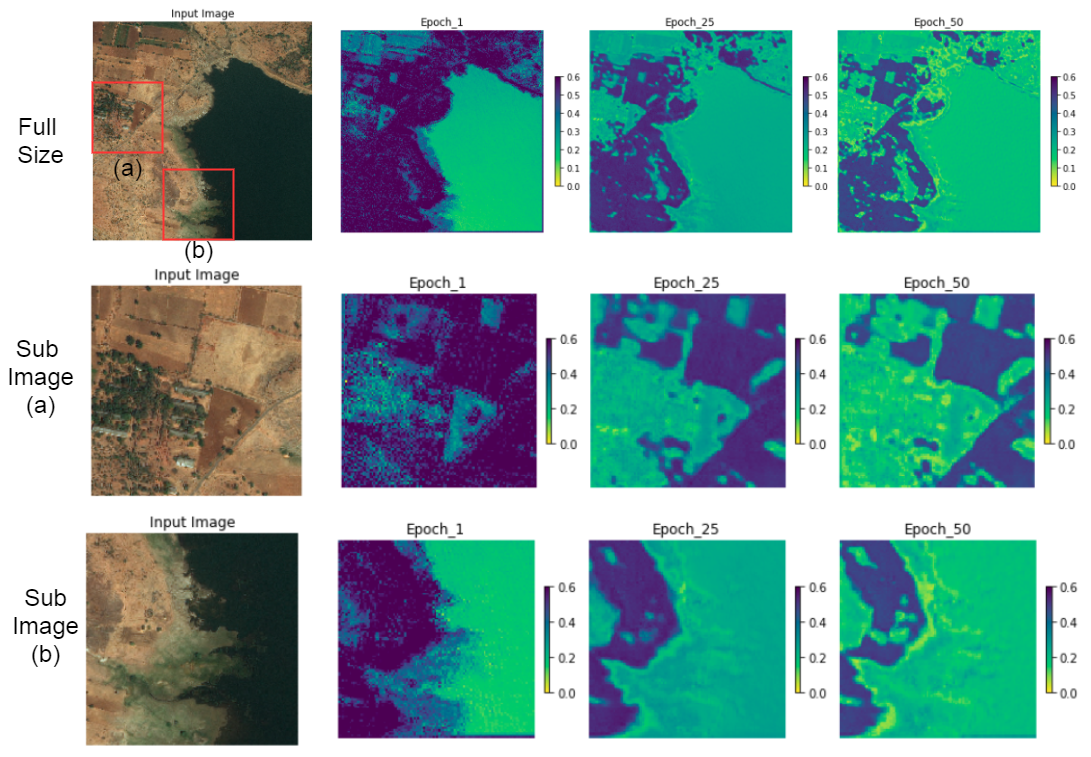}
\caption{Foveation map for one validation image shown in main text Fig.4 at three epoches. Minimum-maximum normalisation applied, where 1 indicting largest FoV-lowest resolution, and 0 indicting smallest FoV-highest resolution.} \label{fig:Time_step_foveation_deepglobe} 
\end{figure}

\clearpage
To show foveation module is effectively learning, we plot foveation map at different training epoches, Fig.~\ref{fig:Time_step_foveation_deepglobe} shows the same example as main text Fig.4 from DeepGlobe dataset and it shows 1) foveation module is progressively refining the learnt spatial distribution of FoV/resolution trade-off; 2) the third row of Fig.~\ref{fig:Time_step_foveation_deepglobe} explains why our foveation approach perform better than GLnet at sub-image (b) as shown in Fig.4, because it shows foveation module learnt to invest small patch at high resolution over coastal areas to capture fine scale context to avoided miss classify it as forest which sharing similar green colour as sea in the input image, as GLnet do.

\clearpage

\subsection{Sensitivity to Hyper-parameters}
\label{result:sensitivity}
\begin{figure}[H]
\centering
\includegraphics[width=0.9\linewidth]{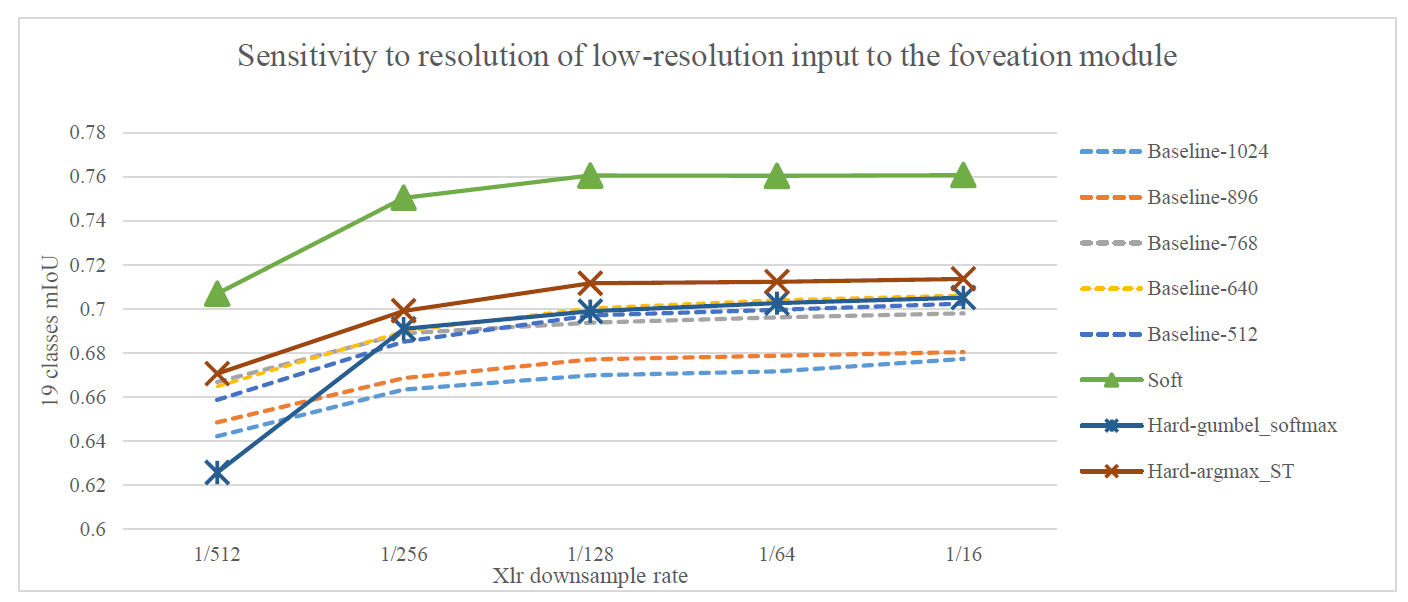}
\caption{\footnotesize Sensitivity study on the resolution of low-resolution input to the foveation module applied during evaluation on Cityscape dataset, 19 classes mIoU evaluated at epoch 100} \label{fig:Sensitivity_study_FXlr_size_cityscape} 
\end{figure}

Here we first evaluate the sensitivity of our approach to the resolution of $F_{\theta}(\mathbf{I}_{\text{lr}})$ at inference time on CityScape dataset, since it forms the most significant computational bottleneck in our approach. Here we apply the same downsampling rate of 1/16 to get $F_{\theta}(\mathbf{I}_{\text{lr}})$ at training time, while at inference time evaluate all models at different downsampling rate to get $F_{\theta}(\mathbf{I}_{\text{lr}})$, with mIoU measured on validation given in Table~\ref{Performance_Cityscape}. In general, Table~\ref{Performance_Cityscape} shows 1) the validation mIoU are increasing with increasing resolution of $F_{\theta}(\mathbf{I}_{\text{lr}})$; 2) ours approaches gain more mIoU boost with increasing resolution of $F_{\theta}(\mathbf{I}_{\text{lr}})$. To better visualise the results, Fig.~\ref{fig:Sensitivity_study_FXlr_size_cityscape} plotted Table~\ref{Performance_Cityscape}, it shows 1) foveated approachs gain more than fixed patch size baselines on mIoU as resolution of $F_{\theta}(\mathbf{I}_{\text{lr}})$ increasing; 2) the gain start converge at downsampling rate of 1/128, at which evaluation time could be reduced to 1/64 comparing to at highest resolution with downsampling rate of 1/16, thus dramatically improve our inference efficiency.

\begin{table}[H]
    
  \centering
  \begin{tabular}{llllll}
    \toprule
    Experiments/$F_{\theta}(\mathbf{I}_{\text{lr}})$ downsample rate & 1/512 & 1/256 & 1/128 & 1/64 & 1/16 \\
    \midrule
    Baseline-$512^2$ & 0.659 & 0.685 & 0.697 & 0.700 & 0.702\\
    Baseline-$640^2$ & 0.665 & 0.690 & 0.700 & 0.704 & 0.706\\
    Baseline-$768^2$ & 0.667 & 0.689 & 0.694 & 0.696 & 0.698\\
    Baseline-$896^2$ & 0.649 & 0.669 & 0.677 & 0.679 & 0.680\\
    Baseline-$1024^2$ & 0.642 & 0.663 & 0.670 & 0.672 & 0.677\\
    \midrule
    Ours-Mean Approximation & {\bfseries0.707} & {\bfseries0.750} & {\bfseries0.761} & {\bfseries0.760} &  {\bfseries0.761}\\
    Ours-GSM & 0.626 & 0.691 & 0.699 & 0.703 & 0.705 \\
    Ours-Mode Approximation & 0.671 & 0.699 & 0.712 & 0.712 & 0.714 \\
    \bottomrule
  \end{tabular}
  \vspace{2mm}
  \caption{\footnotesize Cityscape dataset (HRnet-Backbone), column 2-4: 19 classes mean IoU at epoch 100 evaluated with different resolutions of $\mathbf{x}_{\text{lr}}$; row 2-6: 5 one-hot baselines; row 7-9: our our foveation approach with different sampling approximates.}\label{Performance_Cityscape}

\end{table}

\clearpage
\section{Pseudo-codes}
Here we provide pseudo-codes of our method for the case where each mini-batch is constructed based on a single mega-pixel image (i.e., $L=1$ as described in Sec.~\ref{Architectures_and_Implementation}). We also intend to clean up the whole codebase and release in the final version.

\algdef{SE}[SUBALG]{Indent}{EndIndent}{}{\algorithmicend\ }%
\algtext*{Indent}
\algtext*{EndIndent}

\begin{algorithm}[h]
	\caption{Joint Optimisation of \textit{Foveation Module} and \textit{Segmentation Network}}
	\label{alg:ourmethod}
	\footnotesize
	\begin{algorithmic}
		\State \textbf{Training data}: $\mathcal{D} = \{\textbf{I}_n, \textbf{Y}_n\}_{n=1}^{N}$ 
		\vspace{2mm}
		\State \textbf{Initialize the parameters $\{\theta, \phi\}$ of \textit{foveation module} and \textit{segmentation Network}}
		\vspace{2mm}
		\State \textbf{For $t\in T$ iterations, randomly sample a pair of an images and a label from $\mathcal{D}$:} $\mathbf{I} \in \mathbb{R}^{H\times W \times C}$, $\textbf{Y} \in \mathbb{R}^{H\times W}$
		
		    \Indent
    		\vspace{2mm}
    		\State \textbf{Downsample input images}: $\mathbf{I}_{\text{lr}} \in \mathbb{R}^{h\times w\times C}$
    		\vspace{2mm}
    		\State \textbf{Run \textit{Foveation module}}: $F_{\theta}(\mathbf{I}_{\text{lr}}) \in [0, 1]^{h\times w\times D}$
    		\vspace{2mm}
    		\State \textbf{Select randomly a small subset of pixel locations in $\mathbf{I}_{\text{lr}}$}: $\mathcal{B} \subset \{1,...,wh\}$
    		\vspace{2mm}
    		\State \textbf{For pixel $i\in \mathcal{B}$ in $\mathbf{I}_{\text{lr}}$}:
                \Indent
        	    \vspace{2mm}
        	    \State \textbf{Extract patches from $\mathbf{I}$ at location $i$}:
        	    \vspace{2mm}
        		$$PE(\textbf{I}, i)=\{\mathbf{x}_{1}^{(i)}(\textbf{I}),...,\mathbf{x}^{(i)}_{D}(\textbf{I})\} $$ 
        	    \vspace{-3mm}
        	    
        	    \vspace{2mm}
        	    \State \textbf{Extract the corresponding annotation of the smallest patch $\mathbf{x}_{1}^{(i)}(\textbf{I})$ from $\mathbf{Y}$}:
        	    \vspace{2mm}
        		$$PE^{'}(\textbf{Y}, i)=\{\mathbf{y}_{1}^{(i)}(\textbf{Y})\} $$ 
        	    \vspace{-3mm}
        		
        		\vspace{2mm}
        		\State \textbf{Sample the candidate patch based on \textit{foveation module} output}:
        		$$\mathbf{x}^{(i)} \sim \text{Categorical}\big{(}f^{(i)}_{\theta}(\textbf{I}_\text{lr})\big{)} $$ 
        	    \vspace{-3mm}

        		\State \textbf{Feed the selected patch to \textit{segmentation network}}: 
        		\vspace{-2mm}
        		$$ S_{\phi}\big{(}\mathbf{x}^{(i)}\big{)} $$ 
        	    \vspace{-3mm}
        	    \EndIndent
        	    
    	    \State \textbf{Compute loss and update parameters via a gradient descent}: 
    	    \vspace{-2mm}
    		$$\{\theta, \phi\} \longleftarrow \{\theta, \phi\} - \lambda\cdot \nabla_{\theta, \phi}\Big{[}{1\over{|\mathcal{B}|}}\sum_{i\in \mathcal{B}}\mathcal{L}(S_{\phi}\big{(}\mathbf{x}^{(i)}\big{)}, \mathbf{y}_{1}^{(i)}; \theta, \phi)\Big{]}$$ 
    	    \vspace{-3mm}
		    \EndIndent
	\end{algorithmic}
\end{algorithm}

\end{document}